\documentclass[iicol,pdflatex,sn-apa]{sn-jnl}

\usepackage{graphicx}%
\usepackage{multirow}%
\usepackage{amsmath,amssymb,amsfonts}%
\usepackage{amsthm}%
\usepackage{mathrsfs}%
\usepackage[title]{appendix}%
\usepackage{xcolor}%
\usepackage{textcomp}%
\usepackage{manyfoot}%
\usepackage{booktabs}%
\usepackage{algorithm}%
\usepackage{algorithmicx}%
\usepackage{algpseudocode}%
\usepackage{listings}%
\usepackage{makecell}
\usepackage{placeins}

\theoremstyle{thmstyleone}%
\theoremstyle{thmstyletwo}%

\theoremstyle{thmstylethree}%

\definecolor{orange}{RGB}{0, 0, 0}

\begin{document}

\title[Fix your downsampling ASAP! Aliasing and Sinc Artifact free Pooling in the Fourier domain]{Fix your downsampling ASAP! Aliasing and Sinc Artifact free Pooling in the Fourier domain}


\author[1,2,3]{\fnm{Julia} \sur{Grabinski}}\email{julia.grabinski@uni-mannheim.de}

\author[1]{\fnm{Steffen} \sur{Jung}}\email{steffen.jung@uni-mannheim.de}

\author[1,3]{\fnm{Janis} \sur{Keuper}}\email{janis.keuper@hs-offenburg.de}

\author[1,4]{\fnm{Margret} \sur{Keuper}}\email{margret.keuper@uni-mannheim.de}

\affil[1]{\orgdiv{Department of Machine Learning}, \orgname{University of Mannheim}, \orgaddress{\street{B 6, 26}, \city{Mannheim}, \postcode{68159}, \state{Baden-Württemberg}, \country{Germany}}}

\affil[2]{\orgdiv{Department of High Performance Computing}, \orgname{Fraunhofer Institute for Industrial Mathematics}, \orgaddress{\street{Fraunhofer-Platz 1}, \city{Kaiserslautern}, \postcode{67663}, \state{Rheinland-Pfalz}, \country{Germany}}}

\affil[3]{\orgdiv{Institute for Machine Learning and Analytics}, \orgname{Offenburg University}, \orgaddress{\street{Badstraße. 24}, \city{Offenburg}, \postcode{77652}, \state{Baden-Württemberg}, \country{Germany}}}

\affil[4]{\orgdiv{Department of Computer Vision and Machine Learning}, \orgname{Max Planck Institute for Informatics}, \orgaddress{\street{Saarland Informatcs Campus}, \city{Saarbrücken}, \postcode{66123}, \state{Saarland}, \country{Germany}}}


\abstract{Convolutional Neural Networks (CNNs) are successful in various computer vision tasks.
From an image and signal processing point of view, this success is counter-intuitive, as the inherent spatial pyramid design of most CNNs is apparently violating basic signal processing laws, i.e.~the~\textit{\textbf{Sampling Theorem}} in their downsampling operations.
This issue has been broadly neglected until 
recent work in the context of adversarial attacks and distribution shifts showed that there is a strong correlation between the vulnerability of CNNs and aliasing artifacts induced by bandlimit-violating downsampling. 
As a remedy, we propose an alias-free downsampling operation in the frequency domain, denoted \textit{Frequency Low Cut Pooling (FLC Pooling)} which we further extend to \textit{Aliasing and Sinc Artifact-free Pooling (ASAP)}. ASAP is alias-free and removes further artifacts from sinc-interpolation. 
Our experimental evaluation on ImageNet-1k, ImageNet-C and CIFAR datasets on various CNN architectures demonstrates that networks using FLC~Pooling and ASAP as downsampling methods learn more stable features as measured by their robustness against common corruptions and adversarial attacks, while maintaining a clean accuracy similar to the respective baseline models. }




\maketitle

\section{Introduction}
\begin{figure}[t]%
\centering
\includegraphics[width=\columnwidth]{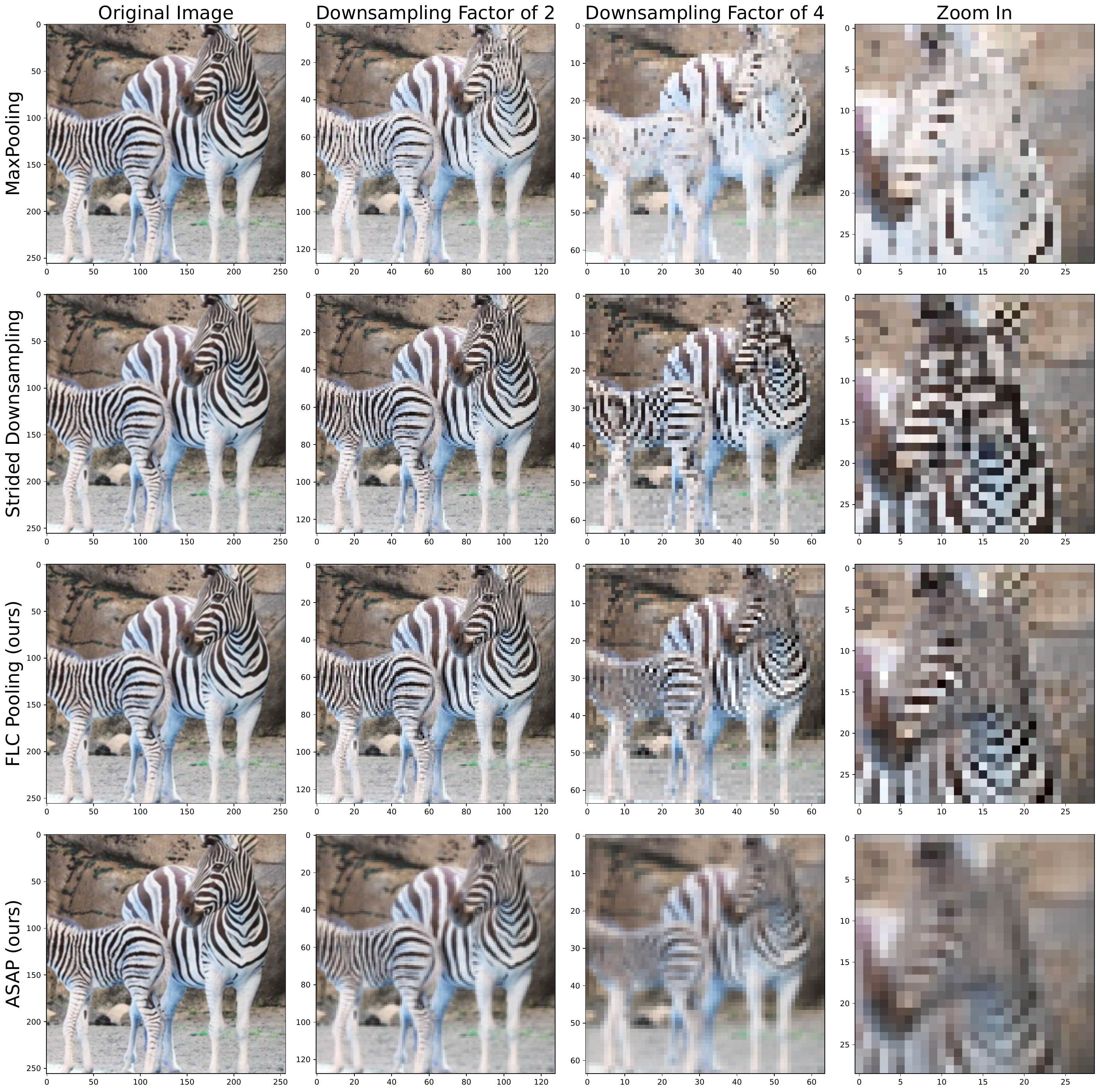}
\caption{The influence of different downsampling methods illustrated with natural images. The first and second rows show the commonly used MaxPooling and strided downsampling. In the third and forth row, we apply our FLC Pooling and aliasing and sinc artifact-free pooling (ASAP), respectively. 
 While MaxPooling does not preserve the image structure well, FLC Pooling and ASAP retain structural and spatial information \textcolor{orange}{much better}. Strided downsampling also preserves, for example, the zebra's structure, yet suffers from severe aliasing artifacts, visible as grid artefacts on the zebra's fur. These artifacts are removed with our FLC Pooling and ASAP. While, FLC Pooling exposes sinc artifacts, visible for example around the zebras head after the first two downsampling stages, such artifacts are removed \textcolor{orange}{with ASAP (here, we show ASAP$_\mathrm{stbl}$, details are given in \autoref{sec:hamming})}.
}\label{fig:teaser_img}
\end{figure}

Most Convolutional Neural Network (CNN) architectures use a combination of small convolutional kernels and downsampling to increase the network's receptive field while keeping the computational costs low. 
However, standard downsampling methods such as MaxPooling, AveragePooling, or Convolution with a stride of two are suffering from a significant drawback: their susceptibility to aliasing \cite{grabinski2022aliasing,zhang2019making,zou2020delving,grabinski2022AAAIw}, which has been shown to correlate with the network's vulnerability to distribution shifts \cite{zhang2019making} and adversarial attacks \cite{grabinski2022aliasing,li2021wavecnet}.  
Based on these results, we investigate the manner in which 2D signals, in case of CNNs' input images and feature maps, are downsampled and how this results in undesired artifacts. So far, prior research mainly focused on aliasing artifacts \cite{grabinski2022aliasing,hossain2023anti,zhang2019making,zou2020delving,grabinski2022AAAIw}, mostly proposing the use of blur kernels for mitigation \cite{zhang2019making,zou2020delving}. However, these approaches are neither capable of removing aliasing completely, nor address   
other types of spectral leakage artifacts related to downsampling in CNNs.

We propose \textbf{F}requency \textbf{L}ow \textbf{C}ut (FLC) Pooling, an aliasing-free method for the downsampling of CNN feature maps in the frequency domain. 
Our experimental evaluation shows that networks using FLC Pooling extract more stable features, as indicated by the models' improved robustness against common corruptions \cite{hendrycks2019robustness}, and adversarial attacks~\cite{harnesssing, pgd}. FLC can even strengthen adversarial training \cite{grabinski2022frequencylowcut,harnesssing} by avoiding catastrophic overfitting.\\
This paper is a consolidation and significant extension of  our previously accepted ECCV~2022~publication~\cite{grabinski2022frequencylowcut}. The original conference paper focused on avoiding catastrophic overfitting during adversarial training by alias-free downsampling. In addition, we here conduct an in-depth study of the properties of CNN feature maps after frequency domain downsampling and propose a further improved method, termed \textbf{A}liasing and \textbf{S}inc \textbf{A}rtifact free \textbf{P}ooling (ASAP). ASAP provides enhanced feature extraction stability, leading to further improved inherent model robustness and increased stability during adversarial training. Particularly, our evaluations show that even aliasing-free pooling methods like FLC Pooling can still be prone to other spectral corruptions, visible as ringing artifacts in the spatial domain.
An example of such artifacts can be observed in \autoref{fig:teaser_img}, specifically in the third row and third column, where the structures near the zebra's head are repeated in a rippled manner.
In contrast, the standard downsampling method used in many CNN architectures, like MaxPooling, completely distorts the zebra's structure (first row \autoref{fig:teaser_img}), and strided convolutions result in severe aliasing artifacts, visible as grid structures on the zebra's fur (second row \autoref{fig:teaser_img}).
Consequently, it is crucial to reevaluate the choice of downsampling techniques employed within CNNs.
In order to reduce artifacts, ASAP consolidates FLC Pooling by using a Hamming window in the frequency domain and considering appropriate padding. 

Our contributions are as follows:
\begin{itemize}
    \item We introduce \textbf{F}requency \textbf{L}ow \textbf{C}ut Pooling, short FLC Pooling, for fully aliasing-free downsampling without additional hyperparameters.
    \item We show that even aliasing-free downsampling can be prone to corruptions in the frequency domain, namely sinc interpolation artifacts.
    \item Consequently, we introduce \textbf{A}liasing and \textbf{S}inc \textbf{A}rtifacts-free \textbf{P}ooling, short ASAP, allowing for more stable feature extraction. 
    \item To validate the robustness of FLC Pooling and ASAP, we empirically evaluate against adversarial attacks \cite{harnesssing,pgd,croce2021mind} as well as common corruptions incorporated in ImageNet-C \cite{hendrycks2019robustness}.
    \item Moreover, we combine FLC Pooling and ASAP with simple FGSM \cite{harnesssing} \textcolor{orange}{and PGD \cite{pgd}} adversarial training and show that the models achieve favorable performance in terms of clean and robust accuracy by avoiding catastrophic overfitting. 
\end{itemize}

\section{Related Work}
\label{sec:related_work}
\subsection{Aliasing in CNNs}
The issue of aliasing effects in CNN-based neural networks has been extensively explored in the literature from various perspectives.
\citet{zhang2019making} enhance the shift-invariance of CNNs by incorporating anti-aliasing filters implemented as convolutions with fixed blur kernels.
Building on this work, shift invariance is further improved in \cite{zou2020delving} by utilizing learned blurring filters instead of predefined kernels.
In \cite{li2021wavecnet}, the pooling operations leverage the low-frequency components of wavelets to mitigate aliasing and enhance robustness against common image corruptions.
 Depth adaptive blurring filters before pooling are proposed in \cite{hossain2023anti}, along with an anti-aliasing activation function.
The importance of anti-aliasing is also recognized in the field of image generation.
The use of blurring filters to eliminate aliases during image generation in generative adversarial networks (GANs) is suggested in \cite{karras2021aliasfree}, while \cite{durall2020watch} and \cite{jung2021spectral} incorporate additional loss terms in the frequency domain to address aliasing.
In \cite{grabinski2022aliasing}, we empirically demonstrate that adversarially robust models exhibit lower levels of aliasing in their downsampling layers compared to non-robust models, using a proposed aliasing measure.
Motivated by these findings, we propose an aliasing-free downsampling method in the frequency domain for stable feature extraction and to prevent catastrophic overfitting.
In extension, we present another approach achieving not only aliasing-free downsampling but also sinc interpolation artifact-free downsampling, which further increases the stability of our extracted features.
\subsection{Spectral Leakage Artifacts}
In contrast to the specific case of aliasing, spectral leakage artifacts have so far received less attention in the context of CNNs.
A common case is the induction of sinc interpolation artifacts, which often arise when applying finite windows to periodic signals.
These artifacts manifest as ringing artifacts in the spatial domain, as described in \cite{gonzales1987digital}, and are associated with the Gibbs phenomenon \cite{hamming1979digital}.
To mitigate these spectral leakage artifacts, various window functions can be employed, as discussed in \cite{prabhu2014window}.

Window functions play a crucial role in the spectral analysis for biomedical image processing \cite{jahne2005digital, semmlow2021biosignal}. 
More recently, spectral leakage artifacts within CNNs have been studied in \cite{tomen2021spectral}, showing that small spatial kernels can contribute to such artifacts. 
Consequently, they propose to learn larger spatial kernels while also applying a Hamming window to the convolution weights. In contrast, we apply window functions in the frequency domain.

\subsection{Robustness}

We assess the feature stability of models relying on two test scenarios, common corruptions and adversarial attacks.

\noindent\textbf{Common Corruptions.}
One aspect to assess the robustness of CNNs involves evaluating their resilience against common corruptions caused by factors such as diverse weather conditions, varying lighting conditions, or subpar camera quality.
To measure this kind of robustness, the widely recognized ImageNet-C dataset is utilized \cite{hendrycks2019robustness}.
This dataset aims to simulate real-world scenarios through synthetic means.
Approaches that improve this form of robustness often employ data augmentation techniques \cite{hendrycks2020augmix, cubuk2018autoaugment}, include shape biasing \cite{geirhos2018imagenet}, or combine adversarial training with augmentations \cite{Kireev2022effectiveness}. \cite{vasconcelos2021impact} use non-trainable lowpass filters to reduce aliasing in the network, increasing the network's robustness against common corruptions. In contrast, our approach completely eliminates aliasing and mitigates sinc-interpolation artifacts, resulting in more stable features and improved robustness against common corruptions.

\noindent\textbf{Adversarial Attacks.}
Adversarial examples are crafted to deceive a network into making incorrect decisions and  expose network-specific vulnerabilities.
In a white-box attack scenario \cite{harnesssing,pgd,croce2021mind}, the attacker has full access to the network's architecture and parameters, while in a black-box attack \cite{andriushchenko2020square}, the attacker only has access to the network's outputs.

One well-known white-box attack is the Fast Gradient Sign Method~\cite{harnesssing}, FGSM, which is an efficient single-step attack. Thus, FGSM is fast to compute, yet not as effective as other methods that use multiple optimization steps, e.g.~as in the white-box Projected Gradient Descent (PGD) \cite{pgd} or in black-box attacks such as Squares \cite{andriushchenko2020square}.  AutoAttack \cite{auto_attack} is an ensemble of different attacks, including an adaptive version of PGD, APGD \cite{croce2021mind}, and is widely used to benchmark adversarial robustness due to its strong performance \cite{robust_bench}.
In this work, we consider adversarial attacks, including FGSM, PGD, APGD and AutoAttack, as a probe of the model's vulnerability. 
We use low-budget, small-epsilon attacks to showcase our increased feature stability without adversarial training.
Additionally, to compare with state-of-the-art adversarial training, we assess performance under strong attacks like AutoAttack and high-epsilon attacks. 


\subsection{Adversarial Training}
Various defense methods have been developed to establish robustness against adversarial attacks.
A key method for improving robustness is adversarial training (AT) \cite{harnesssing, decoupling,wong2020fast}, where networks are exposed to adversarial examples during training.
These adversarial samples are incorporated by introducing an additional loss term during network training \cite{robustness_github, trades}.
Other techniques include utilizing additional training data \cite{carmon2019unlabeled, sehwag2021improving}, particularly the \textit{ddpm} dataset \cite{gowal2021improving, rade2021helperbased, rebuffi2021fixing}, which consists of one million extra samples for CIFAR-10 and is generated using the model proposed by \cite{ho2020denoising}.
Data augmentation has also proven to enhance adversarial robustness \cite{gowal2021uncovering}, and combining it with weight averaging further improves performance \cite{rebuffi2021fixing}.
Some more advanced techniques involve adding specifically generated images to the training dataset~\cite{gowal2021improving}\textcolor{orange}{, additional frequency regularizations \cite{lukasik2023improving} or adding spatial anti-aliasing filters encompassing downsampling layers and activation functions~\cite{rodriguezmunoz2022driver}.}

However, a major drawback of most AT methods is the significant increase in computational resources required for training.
Generating adversaries during training alone can increase training time by a factor of seven to fifteen \cite{pgd, mart, wu2020adversarial, trades}.
Adding additional data or generating new images \cite{gowal2021improving} further amplifies the computational burden. FLC Pooling and ASAP allow to train models efficiently with FGSM adversarial training.

Specifically, FGSM \cite{harnesssing} is a single-step attack and therefore more efficient than more complex, multi-step methods like PGD \cite{pgd}.
In common settings, the iterative process of PGD takes nearly nine times (\autoref{tab:time_report}) longer than FGSM training.
However, FGSM training is susceptible to catastrophic overfitting \cite{wong2020fast}.
Catastrophic overfitting describes the phenomenon during adversarial training (e.g.~with single-step FGSM) where the model overfits to the attack it is trained on, leading to increased susceptibility to other attacks (e.g.~multi-step PGD) \cite{9157154,kim2021understanding}.
While we demonstrate a correlation between aliasing after downsampling and catastrophic overfitting in \cite{grabinski2022aliasing},
in this work we propose downsampling in the frequency domain, specifically FLC Pooling and ASAP, as an option to reduce the risk of catastrophic overfitting. \textcolor{orange}{In consequence, we observe an empirical benefit on efficient adversarial training}.
\section{Aliasing- and Sinc Artifact-Free Pooling}
\label{sec:hamming}
\begin{figure}[t]
	\begin{center}
	\includegraphics[width=0.9\linewidth]{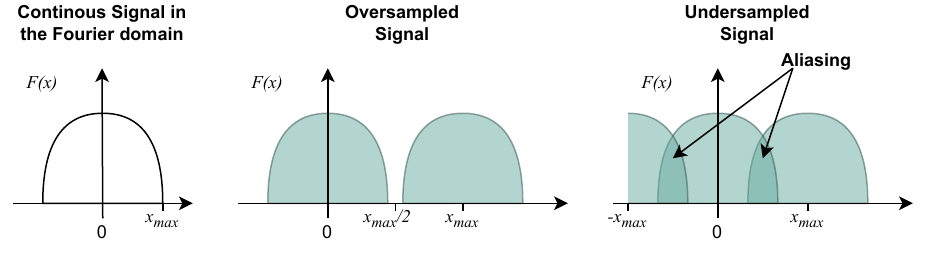}
	\caption[]{Aliasing in the Fourier domain. Left: The frequency spectrum of a 1D signal with maximal frequency $x_{\mathrm{max}}$. After downsampling, replica of the signal appear at a distance proportional to the sampling rate. Center: The spectrum after sampling with a sufficiently large sampling rate. Right: The spectrum after under-sampling with aliases due to overlapping replica.}
	\label{fig:aliasing_theory}
	\end{center}
\end{figure}

First, we present an aliasing-free downsampling method, FLC Pooling, which completely removes aliasing artifacts during downsampling. In contrast, previous approaches \cite{hossain2023anti,zhang2019making,zou2020delving} only reduce aliasing artifacts without fully removing them. Additionally, we further extend FLC
Pooling by introducing Aliasing and Sinc Artifact-free Pooling, short ASAP, which addresses not only the well-known issue of aliasing, but also tackles 
sinc interpolation artifacts.

\vspace{0.2cm}
\noindent{\textbf{Aliasing.}}
Aliasing is a specific type of spectral leakage artifact that occurs when a signal is improperly sampled at a rate below twice the signal's bandwidth \cite{shannon}.
This leads to overlapping high-frequency components, making them indistinguishable from low-frequency components (as visualized in \autoref{fig:aliasing_theory}).
These overlaps in the frequency domain become visible as grid-like artifacts in the spatial domain.

\begin{figure}[t]%
\centering
\includegraphics[width=0.95\columnwidth]{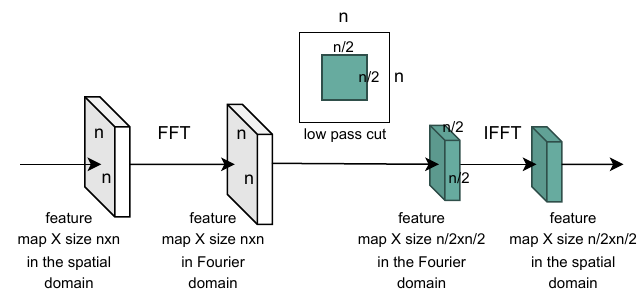}
\caption{Overview of our aliasing-free FLC Pooling. First, we transform the input with the FFT into the frequency domain and shift the low-frequency components in the center. Afterwards, we apply our Frequency Low Cut (FLC) to downsample aliasing-free. Lastly, we transform back into the spatial domain via IFFT.
}\label{fig:method_flc_only}
\end{figure}

\begin{figure*}[t]%
\centering
\includegraphics[width=0.95\textwidth]{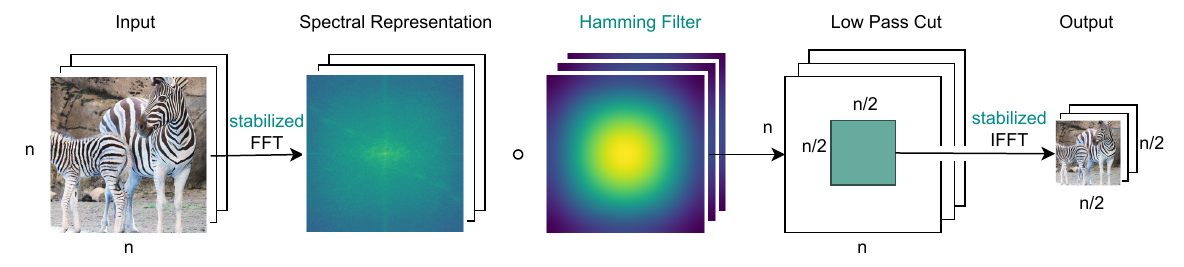}
\caption{Overview over our aliasing and sinc artifact-free downsampling, short ASAP. First, we transform the input with our stabilized FFT into the frequency domain and shift the low-frequency components in the center. Afterwards, we apply a Hamming filter on our frequency representation to prevent sinc interpolation artifacts. Further, we apply our Frequency Low Cut (FLC), which is similar to our FLC Pooling, to downsample aliasing-free. Lastly, we transform back into the spatial domain via our stabilized IFFT.
}\label{fig:method}
\end{figure*}

\vspace{0.2cm}
\noindent\textbf{Aliasing-Free Downsampling.}
Earlier methods reduce aliasing during downsampling via classical blurring operators in the spatial domain \cite{zhang2019making, zou2020delving}.
While those methods reduce aliasing, they can not entirely remove it due to theoretical sampling limitations and limited filter sizes in practice (see \cite{Gonzalez} for details).
In contrast, we perfectly remove aliases in CNNs' downsampling operations without adding additional hyperparameters. We directly address the downsampling operation in the frequency domain, where we can sample according to the Nyquist rate, i.e.~remove all frequencies above $\frac{\mathrm{sampling rate}}{2}$~and thus discard any potential aliases. Our proposed alias-free downsampling operation, FLC Pooling \cite{grabinski2022frequencylowcut}, is visualized in \autoref{fig:method_flc_only}. 

We first perform a Discrete Fourier Transform (DFT) of the feature maps $f$. Feature maps with height $M$ and width $N$ to be downsampled are then represented as
\begin{equation}
\label{fft}
    F(k,l)= \frac{1}{MN} \sum_{m=0}^{M-1} \sum_{n=0}^{N-1} f(m,n)e^{-2\pi j\left(\frac{k}{M}m+\frac{l}{N}n\right)}\,.
\end{equation}
In the resulting frequency space representation $F(k,l)$, all coefficients of frequencies $k,l$, with $|k|$ or $|l|>\frac{\mathrm{sampling rate}}{2}$ have to be set to $0$ before downsampling. 
CNNs commonly downsample with a factor of two, such that the resulting sampling rate is $\frac{1}{2}$. 
Aliasing-free downsampling thus corresponds to removing coefficients where   
$|k|,|l|>\frac{1}{4}$.
The remaining coefficients are then transformed back to the spatial domain via inverse DFT (\autoref{ifft}):
\begin{equation}
\label{ifft}
    f_d(\hat{m}, \hat{n}) =  \frac{1}{\hat{K}\hat{L}} \sum_{k=0}^{\hat{K}-1} \sum_{l=0}^{\hat{L}-1} F_d(k,l)e^{2\pi j\left(\frac{\hat{m}}{\hat{K}}k+\frac{\hat{n}}{\hat{L}}l\right)}.
\end{equation}

\noindent\textbf{Implementation.}
The DFT returns an array $\mathrm{F}$ of complex numbers with size $K\times L$ = $M\times N$, where the frequency $k,l=0$ is stored in the upper left corner 
and the highest frequency is in the center. 
We thus shift the low frequency components into the center of the array via FFT-shift to get $\mathrm{F}_s$  and crop the frequencies below the Nyquist frequency as $\mathrm{F}_{sd}=\mathrm{F}_s[K^\prime:3K^\prime, L^\prime:3L^\prime]$ for $K^\prime=\frac{K}{4}$ and $L^\prime=\frac{L}{4}$, for all samples in a batch and all channels in the feature map. After the inverse FFT-shift, we obtain array $\mathrm{F}_d$ with size $[\hat{K},\hat{L}]=[\frac{K}{2},\frac{L}{2}]$, containing exactly all frequencies below the Nyquist frequency $F_d$, which we can transform back to the spatial domain via inverse DFT for the spatial indices $\hat{m}=0\dots \frac{M}{2}$ and $\hat{n}=0\dots \frac{N}{2}$. 

We receive the aliasing-free downsampled feature map $f_d$ with size $[\frac{M}{2},\frac{N}{2}]$.
Since we are cutting out the low frequencies, we denote this approach \textbf{F}requency \textbf{L}ow \textbf{C}ut Pooling, short FLC Pooling.

\begin{figure}[t]%

\centering

\includegraphics[width=0.95\columnwidth]{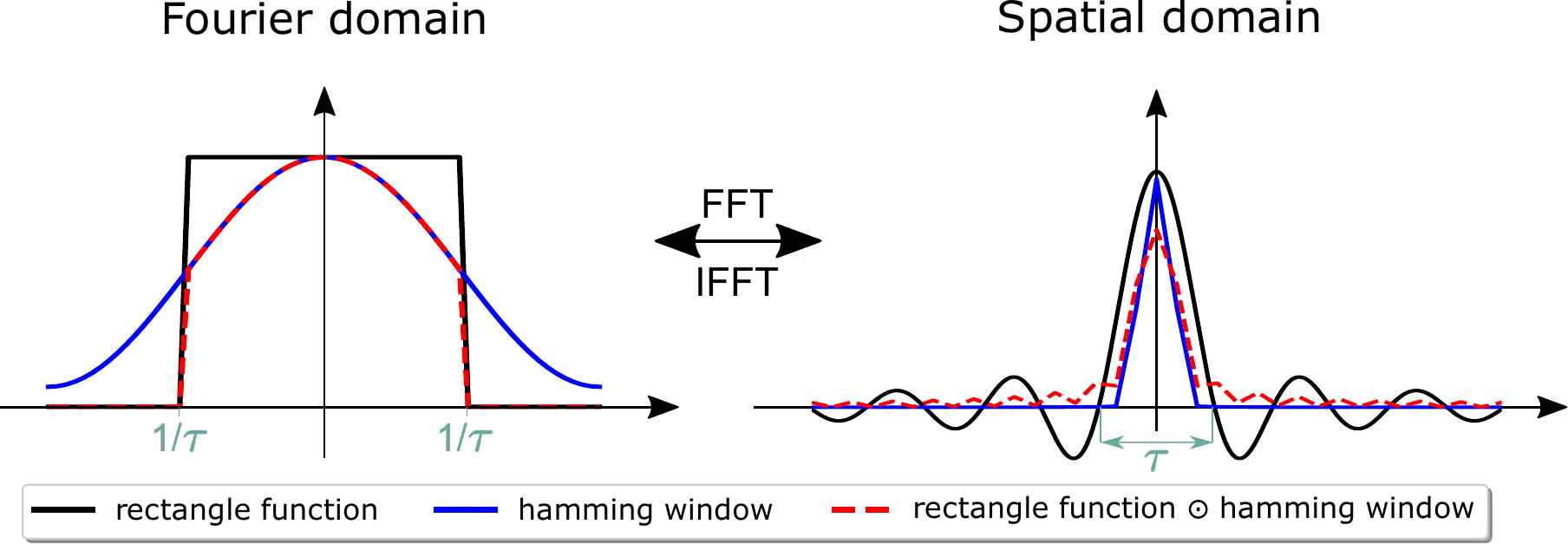} 

\caption{Transformation of the rectangle function (black), the Hamming window (blue) and the point-wise multiplication of those two (red, dashed) from the Fourier domain (left) to the spatial domain (right). The rectangle function transforms to an infinite sinc function with infinite oscillations. In contrast, the Hamming window side lobes become near zero. Thus applying a Hamming window on the rectangle function in the Fourier domain leads to a suppressed version of the sinc in the spatial domain with lower oscillating side lobes.
}\label{fig:rec_sinc}
\end{figure}


\vspace{0.2cm}
\noindent\textbf{Sinc Interpolation Artifacts.} 
\label{sec:sinc_interpolation}
In \textcolor{orange}{\autoref{fig:teaser_img}} we visualize aliasing in the spatial domain.
The leftmost image represents the original image before downsampling, the \textcolor{orange}{other} images depict the downsampled versions (by a factor of two, four and eight in columns two, three and four, respectively).
The second row \textcolor{orange}{(strided downsampling)}, downscaled without any anti-aliasing, exhibits prominent grid artifacts, particularly noticeable in the zebra's fur. 
In contrast, the image downsampled using FLC Pooling in the third row, shows an aliasing-free downscaled version.

However, after examination of the quality and stability of the FLC pooled image, we observe that it is still susceptible to sinc-interpolation artifacts, or \emph{ringing artifacts}, predominantly as also visible around the zebra's head in column four.
In the following, we briefly discuss these artifacts and present an approach to mitigate them.

%
FLC Pooling perfectly removes aliasing artifacts via the low pass cut in the frequency domain.
It implicitly applies a point-wise multiplication with a rectangle function (defined here between frequencies $-1/\tau$ and $1/\tau$): 
\begin{equation}
    \mathit{rect}(\omega, \tau) = \prod \left( \frac{\omega}{2/\tau} \right) =  
    \begin{cases}
      1 & \text{for $|\omega| \leq 1/\tau$} \\
      0 & \text{otherwise}
    \end{cases} 
\end{equation}
with a length of $2/\tau$. Applying a rectangle function in the frequency domain can lead to leakage artifacts (also discussed in~\cite{tomen2021spectral}). Specifically, the equivalent of a rectangle function in the Fourier domain is a sinc function in the spatial domain
\begin{equation}
   f(x, \tau) = F^{-1} \left[ \prod \left( \frac{\omega}{2/\tau} \right) \right] = \frac{1}{\tau\pi} \mathit{sinc} \left(\frac{x}{\tau} \right), 
\end{equation}
as visualized in \autoref{fig:rec_sinc}.

As the rectangle function is applied as point-wise multiplication in the Fourier domain, the equivalent sinc-function in the spatial domain is applied as a circular convolution, i.e.~a convolution with circular boundary conditions. This leads to (i) sinc-interpolation artifacts, also referred to as \textit{ringing artifacts} in the spatial domain, and (ii) to wrap-around effects of the circular sinc-convolution at the boundaries of the feature map, where, for example, signal from the left boundary is convolved into the signal on the right.

\vspace{0.2cm}
\noindent\textbf{Sinc Interpolation Artifact Mitigation.}
To achieve downsampling that is free of both aliasing and sinc interpolation artifacts, it is necessary to apply the rectangle function in the frequency domain. 
There it is common to apply a filter that smoothens the sharp edges of the rectangle function. Therefore, we propose the use of a Hamming window ($H(n)$), which is defined for 1D signals as:
\begin{equation}
    H(n) = \alpha - (1-\alpha)\cdot cos\left(\frac{2n\pi}{N}\right),\quad 0 \leq n \leq N
\end{equation}
The 2D Hamming filter is defined as the outer product of two 1D Hamming filters, where $\alpha = 25/46$ and N represents the number of samples in the signal. However, unlike in \cite{tomen2021spectral}, we do not utilize the Hamming filter as a window function in the spatial domain. Instead, we directly apply the Hamming filter in the frequency domain as a point-wise multiplication. 

The spatial representation of the Hamming window from the frequency domain is shown in \autoref{fig:rec_sinc} (right, blue line). The side slopes of the transformed Hamming window become near zero, effectively reducing interpolation artifacts. Further, we show in \autoref{fig:rec_sinc} the combination of the rectangle function and the Hamming window (red dashed line) in the Fourier domain (left) and the spatial domain (right). The Hamming window facilitates to suppress the oscillations of the sinc function, while remaining completely alias-free. Consequently, the possible artifacts from the circular convolution with the sinc function are reduced and artifacts that would come from the boundary of the feature map are suppressed as the side slope of the signal becomes near zero.

\begin{figure}[t]%
\centering
\includegraphics[width=\columnwidth]{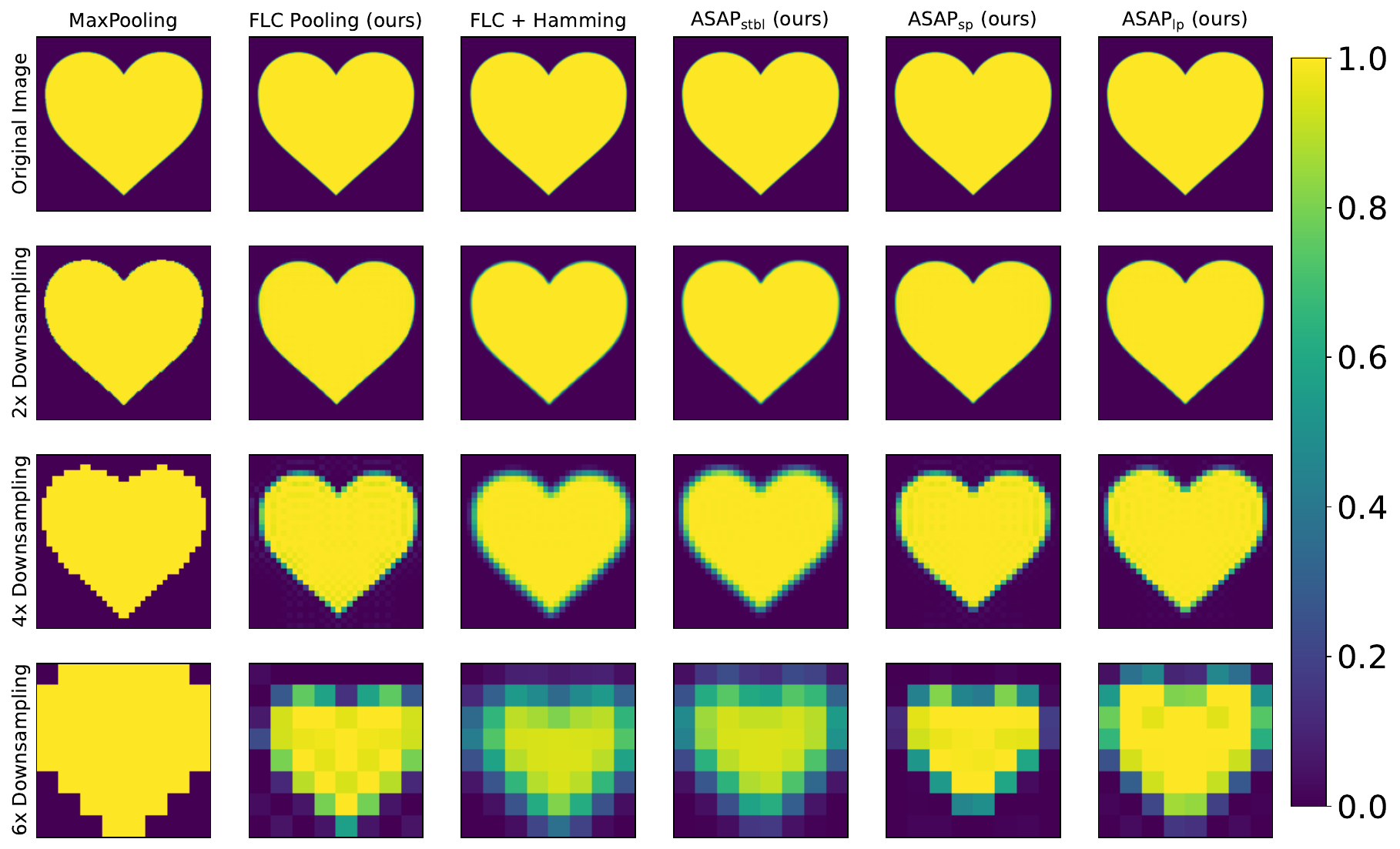}
\caption{We compare the influence of downsampling by different methods. The first \textcolor{orange}{column} shows the commonly used MaxPooling. In the second \textcolor{orange}{column,} we use FLC Pooling, \textcolor{orange}{and in the third column, we show FLC with additional Hamming window applied. I}n the last \textcolor{orange}{three columns,} we show our ASAP \textcolor{orange}{variants, that include the Hamming window and different stabilizations.} 
}\label{fig:heart}
\end{figure}

\vspace{0.2cm}
\noindent\textbf{Stabilized Fast Fourier Transform (FFT).} Further analysis shows that it is beneficial to stabilize the Fourier Transform. Common FFT implementations \cite{cooley1965algorithm} 
 leverage the separability of the Fourier Transform. Typically, the n-dimensional FT are cascaded 1D Fourier transforms. Hence, the 2D FFT is computed first in one direction (vertically) and then in the other direction (horizontally). Due to numerical inaccuracies, this process can introduce shifts in the image after multiple transform applications (see \autoref{fig:heart}, columns two and three).

\begin{figure}[t]%
\centering
\includegraphics[width=0.75\columnwidth]{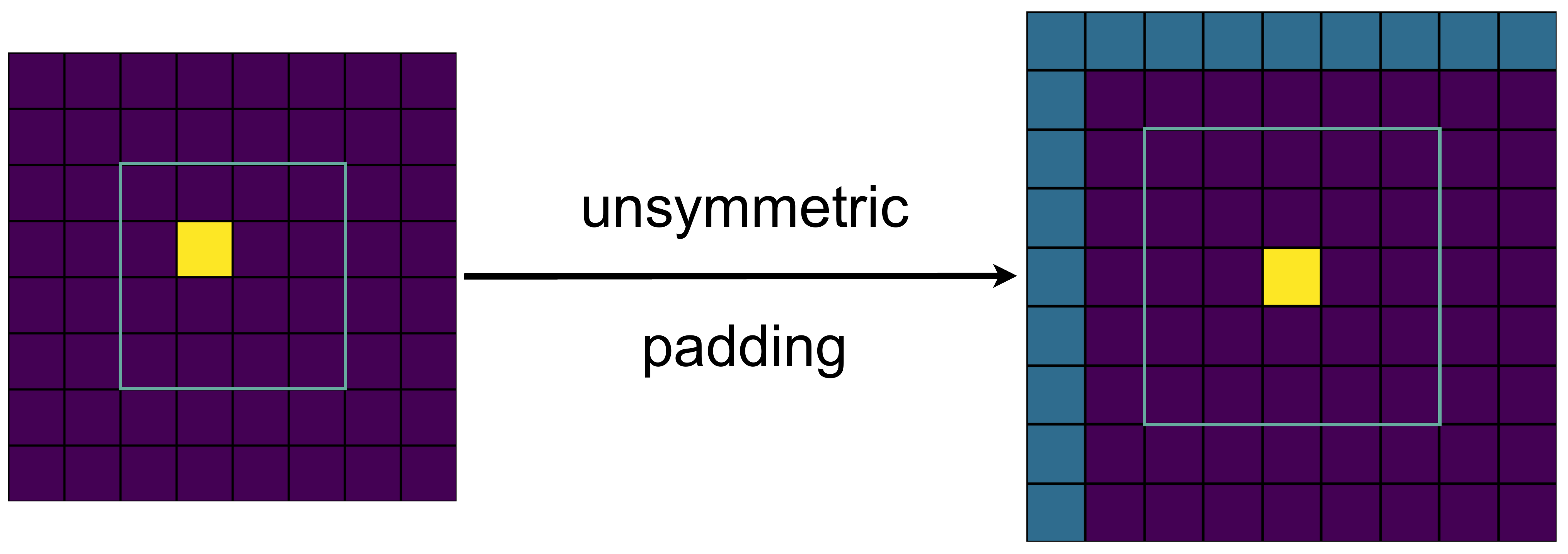}
\caption{We use unsymmetric padding to achieve an uneven input size. Thus, the frequency representation is symmetric and the DC component is centered. During the cut of the low frequency components the cut stays symmetric while this is not the case for the unpadded signal.
}\label{fig:sym_pad}
\end{figure}

We suggest that the numerical issues of the usual row-first FFT originate from the fact that our signal is usually even-sized (even number of pixels in width and height), leading to non-centered frequency representations (see \autoref{fig:sym_pad}, left). The resulting asymmetric cut-out of the low-frequency components can be avoided by padding. Specifically, asymmetric
 padding can center the frequency representation, see \autoref{fig:sym_pad}, right.

\begin{figure}[t]%
\centering
\includegraphics[width=0.7\columnwidth]{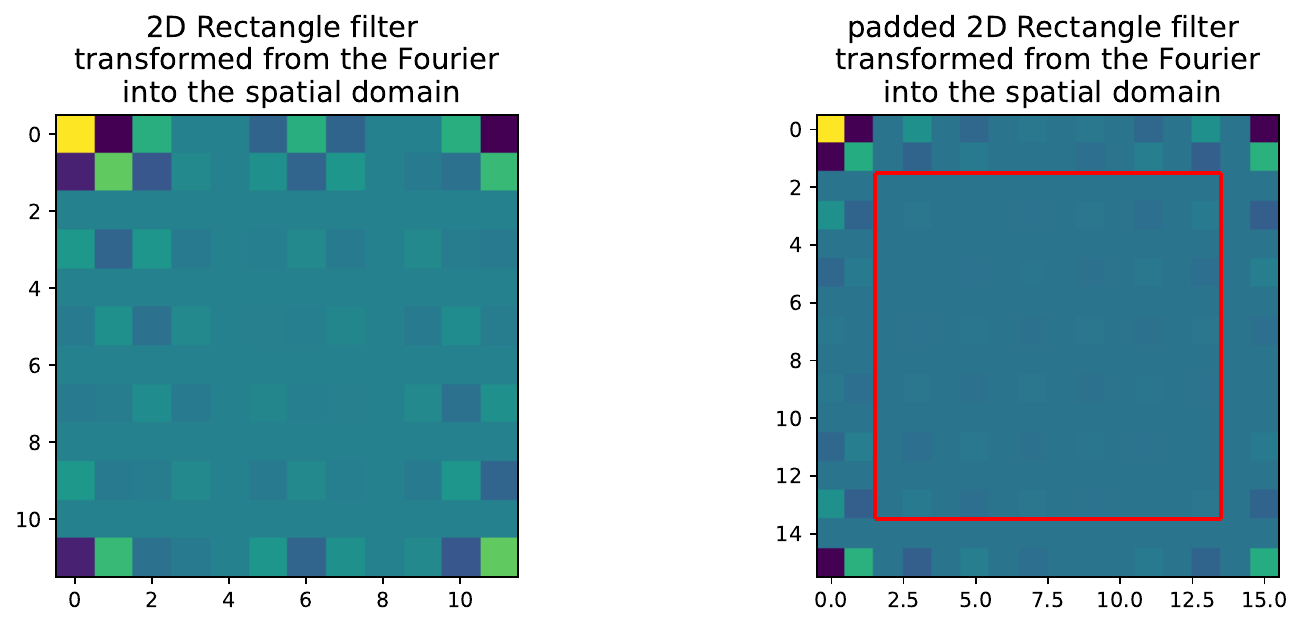}
\caption{Effect of padding (\textcolor{orange}{applied in the} left) against sinc interpolation artifacts. As we only pad for the operation in the frequency domain, the padding is removed afterwards by a centered crop (red line). Thus, some of the dominant sinc interpolation artifacts are removed.
}\label{fig:padding_cut}
\end{figure}

In theory, padding can also serve as a mitigating factor against sinc interpolation artifacts. 
The rationale behind this is twofold. First, the padding compensates for possible ringing artifacts that fold in from the boundaries due to the assumption of periodicity and the application of the rectangular filter. As the padding is removed afterwards these ringing artifacts can thus be removed as shown in \autoref{fig:padding_cut}.
Second, through the padding and transformation into the frequency domain, we artificially increased the resolution of the signal. Thus, when cutting at the Nyquist frequency of the higher resolution signal we artificially increase the value for $1/\tau$ resulting in a more narrow sinc in the spatial domain. \textcolor{orange}{In our analysis, we provide two options for padding. Small padding, denoted by $\textrm{ASAP}_{\textrm{sp}}$, only adds one line of zeros to the bottom and left. Large padding, denoted by $\textrm{ASAP}_{\textrm{lp}}$, pads by $\frac{\mathrm{inputsize}}{2}-1$ to the top and left. Both lead to centered representations (see \autoref{fig:heart}). Larger padding could provide better results while smaller padding is more efficient. Yet, since the FFT algorithm is optimized for data in the array size of powers of two, even small padding can increase the compute costs when the size before padding is exactly a power of two.}

\textcolor{orange}{Therefore, we evaluate an efficient heuristic to avoid the numerical artifacts from the row-first 2D FFTs: we transpose the feature map before every other FFT, such that row-first and column-first FFT are applied in an alternating manner. We denote this stabilization method as $\textrm{ASAP}_{\textrm{stbl}}$. 
We discuss the computation cost} in  \autoref{sec:padding_abliation}.

\section{Experiments}
\label{sec:exp}


For evaluation, we first visualize aliasing artifacts following downsampling, as well as sinc-interpolation artifacts resulting from FLC Pooling in \autoref{subsec:artifact_rep}. 
Second, we assess the performance of state-of-the-art models trained using different downsampling techniques against both our FLC Pooling and ASAP in \autoref{subsec:native_rob}.
This evaluation demonstrates that FLC Pooling and ASAP learn more stable features leading to improved robustness against common corruptions and low budget adversarial attacks.
Third, we evaluate our FLC Pooling and ASAP networks in combination with adversarial training and demonstrate their ability to prevent catastrophic overfitting during FGSM adversarial training in \autoref{subsec:cata_ov}. 
Lastly, we conduct ablation studies in \autoref{sec:abl}.

\begin{figure*}[t]%
\centering
\includegraphics[width=\textwidth]{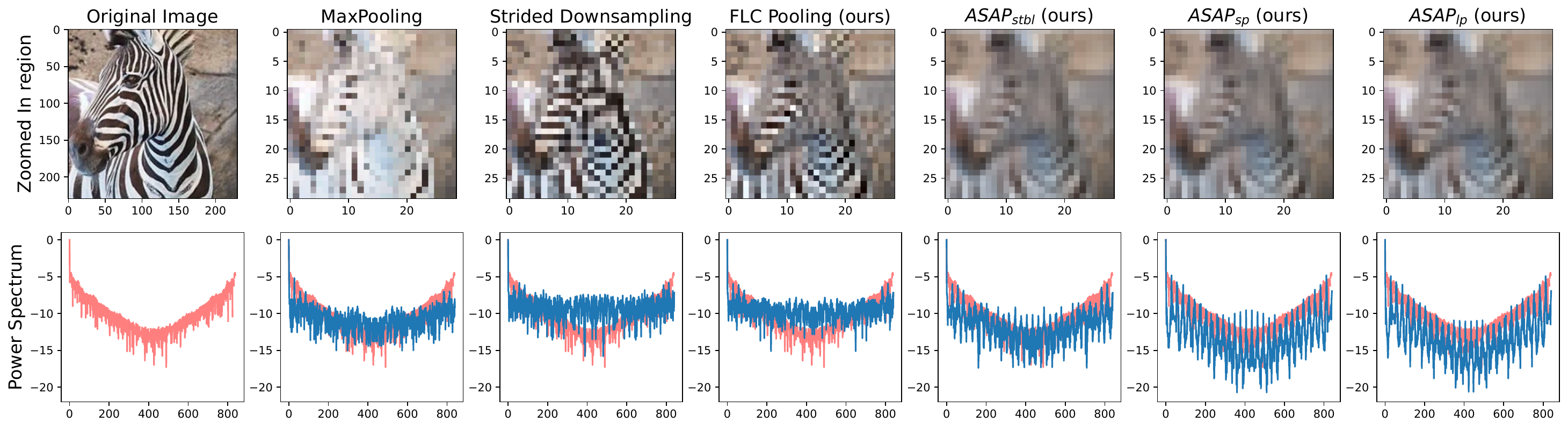}
\caption{Power spectra after different downsampling methods as well as the actual downsampled images. The first row shows the original images and the downsampled versions (downsampled by factor eight) with different pooling methods. After MaxPooling the zebra's fur structures are much less recognizable. When simple downsampling via stride is applied, grid structures appear, and we observe aliasing artifacts. Using FLC Pooling 
removes these aliasing artifacts. However, ringing artifacts surrounding the zebra's head become visible. Only ASAP is able to downsample the image without artifacts. The second row presents the power spectrum (in log scale for the y-axis, x-axis presents the frequency bands) of the images. The first column represents the original power spectrum of the image. Underlying each power spectrum of the downsampled versions we plotted the spectrum of the original image in red. ASAP is the only method to achieve a similar power spectrum to the original image.
}\label{fig:ps_zebra}
\end{figure*}

\subsection{Artifact Representation}
\label{subsec:artifact_rep}
\vspace{0.2cm}
\noindent\textbf{Qualitative Analysis in the Spatial Domain.}
We first visually inspect downsampling artifacts for several downsampling stages (factor 2, 4, and 8) after downsampling by a factor of 8 in a toy example in \autoref{fig:heart}. MaxPooling (first \textcolor{orange}{column}) has the expected effect of disintegrating the spatial structure of the sample. In the bottom row \textcolor{orange}{of column two}, the sinc-interpolation artifacts for FLC Pooling, discussed in \autoref{sec:hamming}, become visible: they appear as ringing artifacts. The same effect can be observed in \autoref{fig:teaser_img} 
and \autoref{fig:ps_zebra}). 
\begin{table}[t]
\caption[]{Mean and standard deviation of the aliasing measure \cite{grabinski2022aliasing} and power spectrum difference measured via  KL divergence after downsampling with conventional and our downsampling methods over 1000 images of CIFAR10. FLC Pooling as well as ASAP do not suffer from aliasing. Hence, aliasing is zero. 
}
\begin{center}
{\begin{tabular}{l@{\,\,}|@{\,\,\,}c@{\,\,\,\,\,}c}
\toprule
Name     &   Aliasing $(\downarrow)$ &  Power Spectrum \\ &&difference $(\downarrow)$\\
\midrule
MaxPooling & 0.26 $\pm$ 0.30 &  0.0113 $\pm$ 0.1019 \\
Strided Conv &  0.17 $\pm$ 0.17 & 0.0025 $\pm$ 0.0070\\
BlurPooling \citeyear{zhang2019making} &  0.17 $\pm$ 0.17 & \textbf{9e-06 $\pm$ 0.0008} \\
ABlurPooling \citeyear{zou2020delving} & 0.22 $\pm$ 0.23 & 0.0007 $\pm$ 0.0115 \\
Wavelet Pooling \citeyear{li2020wavelet} & 0.84 $\pm$ 0.30 & 0.0025 $\pm$ 0.0070 \\
FLC Pooling (ours) & \textbf{0} & 0.0036 $\pm$ 0.0107\\
$\text{ASAP}_{\textcolor{orange}{\textrm{stbl}}}$ (ours) & \textbf{0} & 0.0011 $\pm$ 0.0049 \\
$\textcolor{orange}{\text{ASAP}}_{\textcolor{orange}{\textrm{sp}}}$ \textcolor{orange}{(ours)} & \textcolor{orange}{\textbf{0}} & \textcolor{orange}{0.0006 $\pm$ 0.0075} \\
$\textcolor{orange}{\textrm{ASAP}}_{\textcolor{orange}{\textrm{lp}}}$ \textcolor{orange}{(ours)} & \textcolor{orange}{\textbf{0}} & \textcolor{orange}{0.0004 $\pm$ 0.0071} \\
\bottomrule              
\end{tabular}}
\label{tab:aliasing_ps}
\end{center}
\end{table}
%
%
%
To mitigate these artifacts, we apply a Hamming filter, as described in \autoref{sec:hamming}, suppressing these artifacts 
(\textcolor{orange}{four last columns} of \autoref{fig:heart}). A remaining, potentially undesired effect is the slight shift of the signal to the lower right, which is removed by the stabilization in the full $\text{ASAP}$ \textcolor{orange}{variants}. 

\vspace{0.2cm}
\noindent\textbf{Analysis in the Frequency Domain.}
\autoref{fig:ps_zebra} depicts the 1D power spectrum after downsampling for a more realistic example.
The first column shows the original image with the full power spectrum.
Each column presents a different downsampling technique, with the qualitative result after downsampling in the first row and its power spectrum in the second row.
In contrast, \textcolor{orange}{all our} ASAP \textcolor{orange}{variants} obtain a power spectrum similar to the one of the original image.

\vspace{0.2cm}
\noindent\textbf{Quantitative Analysis.}
To quantify, we evaluate the aliasing measure proposed in \cite{grabinski2022aliasing} as well as the difference in power spectrum for  different downsampling methods 
in \autoref{tab:aliasing_ps}. Methods based on blurring before downsampling, including \textcolor{orange}{all variants of our}  proposed ASAP, maintain the power spectrum of the original image best. 
However, our FLC Pooling and \textcolor{orange}{all variants of our}  ASAP are the only entirely alias-free approaches. $\text{ASAP}_{\textcolor{orange}{\textrm{lp}}}$ offers \textcolor{orange}{the most} favorable trade-off, being alias-free and preserving the power spectrum well.


\begin{table}[t]
\caption[]{Time evaluation of \textcolor{orange}{our} additional padding \textcolor{orange}{compared to no padding}. We evaluate the time for one execution on $32\times32$ CIFAR10 input images in 100 independent runs over the validation set on an NVIDIA A100. Small padding only adds one line of zeros on the bottom and left. \textcolor{orange}{Large padding pads by $\frac{\mathrm{inputsize}}{2}-1$ to the top and left}. A sequence includes three operations stacked, such that the input is downsampled by a factor of eight, while a single execution only downsamples by a factor of two.
}
\centering
\begin{tabular}{l|cc}
\toprule
Padding &Sec Per & Sec Per \\
 Size   &   Single Execution & Sequence\\
\midrule
No Padding & 0.0522 $\pm$ 0.0181 & 0.0738 $\pm$ 0.0195 \\
Small Padding & 0.0623 $\pm$ 0.0211 & 0.0794 $\pm$ 0.0185\\ 
Large Padding & 0.2903 $\pm$ 0.0254 & 0.3782 $\pm$ 0.0282 \\ 
\bottomrule              
\end{tabular}
\label{tab:pad_time}
\end{table}

\vspace{0.2cm}
\noindent\textbf{FFT stabilization.} 
Following \autoref{sec:hamming}, we apply the FFT by using our stabilization heuristic or additional padding.
\autoref{fig:heart} (columns four to six) depict the effect of using the stabilization heuristic by transposing ($\textrm{ASAP}_{\textrm{stbl}}$), large $\textrm{ASAP}_{\textrm{lp}}$ and small $\textrm{ASAP}_{\textrm{sp}}$ asymmetric padding before transforming into the frequency domain so that all representations are correctly centered. 
As additional padding potentially increases the computational cost of the FFT, leading to an increase in computational costs for $\textrm{ASAP}_{\textrm{sp}}$ by $8$\% and by a factor of $5.6$ for $\textrm{ASAP}_{\textrm{lp}}$ on an NVIDIA A100 as reported in \autoref{tab:pad_time}.

\begin{table*}[t]
\caption[]{Clean and robust accuracy \textcolor{orange}{(in percent)} and performance under common corruptions for several \textcolor{orange}{common} models trained without adversarial training on ImageNet-1k. Attacks are \textcolor{orange}{performed} with $\epsilon=1/255$ \textcolor{orange}{to probe the stability of learned representations. While robustness of adversarially trained models is typically tested with $\epsilon=4/255$, this attack strength would cause all networks to fail completely.} FLC Pooling and ASAP improve prediction stability under adversarial attacks. In addition, $\textrm{ASAP}_{\textcolor{orange}{\textrm{sp}}}$  outperforms all methods on common corruptions.
}
\footnotesize
\begin{center}
\begin{tabular}{c|l|cccccc}
\toprule
Arch & Method     &   Acc@1 & Acc@5  &  APGD  &   FGSM  &Corr@1& Corr@5\\ 
\toprule
\multirow{7}{*}{\rotatebox[origin=c]{90}{ResNet-18}}
& Baseline & 69.56 &89.09 &0.01&21.20&34.37&       54.66  \\
&BlurPooling \citeyear{zhang2019making} &  71.38 & 90.12  & 0.07 & 21.78  &   35.97 &        56.48\\
&Wavelet Pooling \citeyear{li2020wavelet} & 71.29 & 90.12 & 0.01 & 23.10 & 37.53 &   58.34 \\
&FLC Pooling (ours) & 69.16 & 88.91 & \textbf{0.32} & 35.21 & 40.19 &     61.82\\
&$\textrm{ASAP}_{\textcolor{orange}{\textrm{stbl}}}$ (ours) & 69.53 & 89.11 & 0.31 & 36.44 & 40.33 &   61.99 \\
&$\textrm{ASAP}_{\textcolor{orange}{\textrm{lp}}}$ (ours) & 71.18 & 89.93 & 0.31& \textbf{40.33} &  43.14 &   64.63\\
&$\textrm{ASAP}_{\textcolor{orange}{\textrm{sp}}}$  (ours) & \textbf{71.54} & \textbf{90.24} &  0.28 &  39.57 & \textbf{43.44} &    \textbf{64.87} \\
\midrule
\multirow{10}{*}{\rotatebox[origin=c]{90}{ResNet-50}}
&Baseline & 75.85 & 92.88 &  0.07   &  36.00   & 40.80   &     61.18 \\ 
&BlurPooling \citeyear{zhang2019making} &  77.19 & 93.38 &0.24 & 39.44 &  43.03 &    63.60  \\
 &Wavelet Pooling \citeyear{li2020wavelet} & 76.56 & 92.95 & 0.13 &38.45 & 42.13&   62.79 \\
& \textcolor{orange}{low-pass \citeyear{vasconcelos2021impact}} & \textcolor{orange}{77.50} & \textcolor{orange}{-} & \textcolor{orange}{-} & \textcolor{orange}{-}  & \textcolor{orange}{30.00} & \textcolor{orange}{-} \\
& \textcolor{orange}{low-pass \citeyear{vasconcelos2021impact} + RA } &  \textcolor{orange}{78.40} & \textcolor{orange}{-} & \textcolor{orange}{-} & \textcolor{orange}{-}  & \textcolor{orange}{34.50} & \textcolor{orange}{-} \\
& \textcolor{orange}{low-pass \citeyear{vasconcelos2021impact} + Swish + RA } & \textcolor{orange}{ \textbf{78.80}} & \textcolor{orange}{-} & \textcolor{orange}{-} & \textcolor{orange}{-}  &  \textcolor{orange}{35.10} & \textcolor{orange}{-} \\
 &FLC Pooling (ours) &77.13 & 93.44& 0.53 &\textbf{59.05} & 50.51    &71.59\\ 
&$\textrm{ASAP}_{\textcolor{orange}{\textrm{stbl}}}$ (ours) &77.12& 93.45& 0.67 & 57.72 & 50.76 &    71.87 \\ 
&$\textrm{ASAP}_{\textcolor{orange}{\textrm{lp}}}$ (ours) & 78.11 & 94.00 & 0.67 & 58.37 & 53.34 & 74.00 \\
&$\textrm{ASAP}_{\textcolor{orange}{\textrm{sp}}}$ (ours) & 78.54 & \textbf{94.10} & \textbf{0.94} & 56.84 &  \textbf{54.20} & \textbf{74.83} \\
\midrule
\multirow{7}{*}{\rotatebox[origin=c]{90}{ResNet-101}}
&Baseline& 77.25 & 93.54&0.04 &38.66& 46.09 &   67.01  \\
&BlurPooling \citeyear{zhang2019making} & 78.15 & 94.03 & 0.25& 42.76 &  46.92 &     67.70 \\
&Wavelet Pooling \citeyear{li2020wavelet} & 78.06 & 93.96 & 0.04 & 46.11& 48.49 & 69.16 \\
&FLC Pooling (ours) &  78.16 & 93.95 & 1.26 & 58.49 & 52.91 &      73.72 \\
&$\textrm{ASAP}_{\textcolor{orange}{\textrm{stbl}}}$ (ours) & 78.11 & 94.12 & 1.32 & \textbf{59.56} & 53.13 & 73.99 \\
&$\textrm{ASAP}_{\textcolor{orange}{\textrm{lp}}}$  (ours) & 79.07  & 94.35 & 1.57 & 58.99 & 55.82 & 76.19 \\
&$\textrm{ASAP}_{\textcolor{orange}{\textrm{sp}}}$ (ours) & \textbf{79.34} & \textbf{94.63} &\textbf{1.78 }& 58.06 & \textbf{56.14} &  \textbf{76.51 } \\
\midrule
\multirow{7}{*}{\rotatebox[origin=l]{90}{WRN-50-2}}
& Baseline & 78.29 &94.03 & 0.28 &37.74 & 45.23 &     65.75 \\
&BlurPooling \citeyear{zhang2019making} & 78.60 & 94.18 & 0.60 &39.39 &  46.22 &     66.58 \\
&FLC Pooling (ours) & 79.67 & 94.74 & \textbf{0.64}  & \textbf{54.00} & 48.48 &      69.22\\
 & $\textrm{ASAP}_{\textcolor{orange}{\textrm{stbl}}}$ (ours) &79.68 & 94.71 & 0.61 &  53.43 & 48.99 &     69.63 \\
&$\textrm{ASAP}_{\textcolor{orange}{\textrm{lp}}}$  (ours) & 80.01 & \textbf{94.98} & 0.45 & 49.88 & 50.17 &   70.40 \\
&$\textrm{ASAP}_{\textcolor{orange}{\textrm{sp}}}$ (ours) &  \textbf{80.31} & 94.96  & 0.51 & 48.52 &  \textbf{50.93} &  \textbf{70.98}  \\
\midrule
\multirow{7}{*}{\rotatebox[origin=c]{90}{MobileNet-v2}}
&Baseline & 71.36  &  90.12 & 0.00 &14.29 & 34.13 &        54.43\\ 
&BlurPooling \citeyear{zhang2019making} &  \textbf{72.47} & \textbf{90.69}  & 0.00& 13.37 & 34.33 &  54.50\\
&Wavelet Pooling \citeyear{li2020wavelet} & 71.94 & 90.46 & 0.00 &13.04 & 34.28 &       54.61\\
 & FLC Pooling (ours) &66.81 & 87.72 &0.26& 25.21 & 34.70 &       55.64\\ 
&$\textrm{ASAP}_{\textcolor{orange}{\textrm{stbl}}}$ (ours) &66.83 &87.71& 0.29 & 25.58 & 35.10 &   56.18 \\ 
&$\textrm{ASAP}_{\textcolor{orange}{\textrm{lp}}}$ (ours) &  68.70 & 88.80 & \textbf{0.42} & \textbf{28.37}& 37.15 &  58.48 \\
&$\textrm{ASAP}_{\textcolor{orange}{\textrm{sp}}}$ (ours) & 69.14 & 88.87 & 0.41 &  28.06 &  \textbf{38.34} &    \textbf{59.98} \\
\bottomrule      
\end{tabular}
\label{tab:imagenet_acc}
\end{center}
\end{table*}
\begin{table*}[t]
\caption[]{Mean accuracy \textcolor{orange}{(in percentage)} and standard deviation on clean samples, perturbed samples with FGSM~\cite{harnesssing} and PGD~\cite{pgd} as well as corrupted samples~\cite{hendrycks2019robustness} for \textcolor{orange}{four different architectures} (five different random seeds) trained without adversarial training on CIFAR-10. \textcolor{orange}{Attacks are done with an epsilon of $\epsilon=1/255$ and corruption performance is reported as mean over all severities.} 
For CIFAR-10 our ASAP and FLC Pooling outperform the baseline and show overall a high robustness against adversarial attacks and common corruptions. 
}
\footnotesize
\begin{center}
\begin{tabular}{c|l|cccc}
\toprule
Arch & Method     &   Acc@1 &  FGSM  &   PGD & Corruptions\\
\toprule
\multirow{7}{*}{\rotatebox[origin=c]{90}{ResNet-9}}
&\textcolor{orange}{Baseline \citeyear{lukasik2023improving}} & \textcolor{orange}{94.29} & \textcolor{orange}{59.58} & \textcolor{orange}{53.04} & \textcolor{orange}{-} \\ 
&\textcolor{orange}{DCT Conv WD \citeyear{lukasik2023improving}} & \textcolor{orange}{93.18} & \textcolor{orange}{59.25} & \textcolor{orange}{56.08} &\textcolor{orange}{-} \\
& \textcolor{orange}{DCT Conv SD \citeyear{lukasik2023improving}} & \textcolor{orange}{93.09} & \textcolor{orange}{59.87} & \textcolor{orange}{56.89} & \textcolor{orange}{-}\\

 & \textcolor{orange}{FLC Pooling (ours)} 
& \textcolor{orange}{94.53 $\pm$ 0.11}  & \textcolor{orange}{\textbf{69.05 $\pm$ 0.2}2}  & \textcolor{orange}{\textbf{66.60 $\pm$ 0.58}} & \textcolor{orange}{\textbf{75.29  $\pm$    0.75}} \\
&$\textcolor{orange}{\textrm{ASAP}}_{\textcolor{orange}{\textrm{stbl}}}$ \textcolor{orange}{(ours)} & \textcolor{orange}{\textbf{94.56 $\pm$ 0.16}} & \textcolor{orange}{68.96 $\pm$ 0.61} & \textcolor{orange}{65.81 $\pm$ 0.89} & \textcolor{orange}{74.79 $\pm$ 0.64}\\ 
& $\textcolor{orange}{\textrm{ASAP}}_{\textcolor{orange}{\textrm{lp}}}$ \textcolor{orange}{(ours)} & \textcolor{orange}{94.52 $\pm$ 0.15} & \textcolor{orange}{68.56 $\pm$ 0.52}  & \textcolor{orange}{66.06 $\pm$ 1.04} &\textcolor{orange}{75.04 $\pm$    0.73}  \\ 
& $\textcolor{orange}{\textrm{ASAP}}_{\textcolor{orange}{\textrm{sp}}}$ \textcolor{orange}{(ours)} & \textcolor{orange}{94.55 $\pm$ 0.08}  & \textcolor{orange}{68.52 $\pm$ 0.53}  & \textcolor{orange}{65.28 $\pm$ 1.03} & \textcolor{orange}{74.61  $\pm$   0.73} \\
\midrule
\multirow{10}{*}{\rotatebox[origin=c]{90}{ResNet-18}}
& Baseline &  93.03 $\pm$ 0.13 & 78.62 $\pm$ 0.28 & 72.49 $\pm$ 0.67  & 76.93  $\pm$ 0.45    \\
&BlurPooling \citeyear{zhang2019making} &  \textbf{93.25 $\pm$ 0.17} & 79.24 $\pm$  0.32 & 75.23 $\pm$ 0.55 & 77.70 $\pm$ 0.54\\
&ABlurPooling \citeyear{zou2020delving} & 92.77 $\pm$ 0.15 & \textbf{79.65 $\pm$ 0.52} & \textbf{76.94 $\pm$ 0.79} & 76.59 $\pm$ 0.33 \\
 &WaveletPooling \citeyear{li2020wavelet} & 93.00 $\pm$ 0.06 & 79.15 $\pm$ 0.15 & 72.71 $\pm$ 0.55 & 78.40 $\pm$ 0.23\\
&\textcolor{orange}{DCT Conv WD \citeyear{lukasik2023improving}} & \textcolor{orange}{88.80 $\pm$     0.16} & \textcolor{orange}{60.53 $\pm$   1.06} & \textcolor{orange}{58.65 $\pm$   1.21} & \textcolor{orange}{73.88 $\pm$  0.41} \\
& \textcolor{orange}{DCT Conv SD \citeyear{lukasik2023improving}} & \textcolor{orange}{89.93 $\pm$   0.10} & \textcolor{orange}{61.37 $\pm$ 0.51} & \textcolor{orange}{59.40   $\pm$  0.74} & \textcolor{orange}{75.84 $\pm$ 0.35} \\
&FLC Pooling (ours) 
& 93.12 $\pm$ 0.19 & 78.92 $\pm$ 0.26 & 74.17 $\pm$ 0.60 &78.59 $\pm$ 0.29  \\
&$\textrm{ASAP}_{\textcolor{orange}{\textrm{stbl}}}$ (ours) & 93.12 $\pm$ 0.25 & 79.08 $\pm$ 0.43 & 75.06 $\pm$ 0.76 & \textbf{78.68 $\pm$ 0.19}\\
&$\textrm{ASAP}_{\textcolor{orange}{\textrm{lp}}}$ (ours) & 93.24 $\pm$ 0.15 & 79.17 $\pm$ 0.23 & 74.94 $\pm$ 0.56 & 78.65 $\pm$ 0.33\\
&$\textrm{ASAP}_{\textcolor{orange}{\textrm{sp}}}$ (ours) &  93.00 $\pm$ 0.12 & 79.12 $\pm$ 0.49 & 74.69 $\pm$ 1.36 & 78.42 $\pm$ 0.20 \\
\midrule
\multirow{8}{*}{\rotatebox[origin=c]{90}{WRN-50-2}}
&Baseline &94.33 $\pm$ 0.13& 77.92 $\pm$ 0.64& 69.36 $\pm$ 1.20 & 77.08 $\pm$ 0.38 \\
&BlurPooling \citeyear{zhang2019making} & 94.42 $\pm$ 0.12 & 76.21 $\pm$ 0.30 & 68.66 $\pm$ 0.49 & 77.59 $\pm$ 0.47\\
&ABlurPooling \citeyear{zou2020delving} & 93.66 $\pm$ 0.18 & 78.26 $\pm$ 1.50 & 71.76 $\pm$ 1.91 & 76.74 $\pm$ 1.10\\
&WaveletPooling \citeyear{li2020wavelet} & 94.44 $\pm$ 0.13 & 78.12 $\pm$ 1.11 & 69.26 $\pm$ 1.02 & \textbf{79.95 $\pm$ 0.42} \\
&FLC Pooling (ours)
& 94.33 $\pm$ 0.20 & 75.41 $\pm$ 0.41 & 66.30 $\pm$ 0.87  & 79.33 $\pm$ 0.43\\
&$\textrm{ASAP}_{\textcolor{orange}{\textrm{stbl}}}$(ours) & \textbf{94.51 $\pm$ 0.17} & 77.22 $\pm$ 0.89 & 71.24 $\pm$ 1.86 & 79.90 $\pm$ 0.37\\
&$\textrm{ASAP}_{\textcolor{orange}{\textrm{lp}}}$ (ours) & 94.20 $\pm$ 0.18 & \textbf{78.63 $\pm$ 0.57} & \textbf{72.36 $\pm$ 0.67} & 79.72 $\pm$ 0.41\\
&$\textrm{ASAP}_{\textcolor{orange}{\textrm{sp}}}$(ours) &  94.16 $\pm$ 0.21 & 77.30 $\pm$ 0.24 & 70.74 $\pm$ 1.14 & 79.61 $\pm$ 0.82 \\
\midrule
\multirow{5}{*}{\rotatebox[origin=c]{90}{AlexNet}}
&\textcolor{orange}{Baseline} & \textcolor{orange}{\textbf{89.45 $\pm$ 0.23}} & \textcolor{orange}{69.28 $\pm$ 0.22} & \textcolor{orange}{69.97 $\pm$ 0.28} & \textcolor{orange}{73.74 $\pm$ 0.13}   \\
&\textcolor{orange}{FLC Pooling (ours)} & \textcolor{orange}{87.80 $\pm$ 0.10} & \textcolor{orange}{70.40 $\pm$ 0.36} & \textcolor{orange}{71.49 $\pm$ 0.36} & \textcolor{orange}{\textbf{74.33 $\pm$  0.50}}   \\
 &$\textcolor{orange}{\textrm{ASAP}}_{\textcolor{orange}{\textrm{stbl}}}$ \textcolor{orange}{(ours)} &\textcolor{orange}{87.59 $\pm$ 0.23} & \textcolor{orange}{70.50 $\pm$ 0.26} & \textcolor{orange}{71.52 $\pm$ 0.22} & \textcolor{orange}{73.84 $\pm$ 0.27}  \\
&$\textcolor{orange}{\textrm{ASAP}}_{\textcolor{orange}{\textrm{lp}}}$ \textcolor{orange}{(ours)} &  \textcolor{orange}{87.90 $\pm$ 0.09} & \textcolor{orange}{\textbf{70.87 $\pm$ 0.26}} & \textcolor{orange}{\textbf{71.93 $\pm$ 0.20}} & \textcolor{orange}{74.02 $\pm$ 0.42}\\
&$\textcolor{orange}{\textrm{ASAP}}_{\textcolor{orange}{\textrm{sp}}}$ \textcolor{orange}{(ours)}& \textcolor{orange}{87.68 $\pm$ 0.16} & \textcolor{orange}{70.71 $\pm$ 0.30} & \textcolor{orange}{71.82 $\pm$ 0.26} & \textcolor{orange}{73.94 $\pm$ 0.21} \\
\bottomrule              
\end{tabular}
\label{tab:cleantrain_cifar_acc}
\end{center}
\end{table*}
\subsection{Feature Stability without Adversarial Training}
\label{subsec:native_rob}

We evaluate the stability of the features learned by our proposed FLC Pooling and ASAP by considering different adversarial samples as well as common corruptions \cite{hendrycks2019robustness}. When models are trained without adversarial training, using low $\epsilon$ budgets makes most sense \cite{lukasik2023improving}. These models can not be expected to be robust against strong attacks but we merely want to probe the improved stability of their learned representation. We conduct experiments using two different datasets, ImageNet-1k \cite{imagenet} and CIFAR-10 \cite{krizhevsky2009learning}. 

\vspace{0.2cm}
\noindent\textbf{For high-resolution data}, we use ImageNet-1k \cite{imagenet} and trained one network per ASAP method (stabilized FFT, large and small padding). The baseline networks utilized the pre-trained weights provided by PyTorch.
The weights for BlurPooling are provided by \cite{zhang2019making} and for Wavelet Pooling by \cite{li2020wavelet}.
\cite{zou2020delving} only provide weights for ResNet-101.
For our FLC Pooling and ASAP, we follow the training procedures suggested by the original authors of each network.

\autoref{tab:imagenet_acc} shows the performance of each network on clean, perturbed, and corrupted versions of the ImageNet-1k dataset.
We observe that all models benefit from the use of our ASAP method for the robustness against common corruptions.
Consistently, $\textrm{ASAP}_{\textcolor{orange}{\textrm{sp}}}$  outperforms all other methods.
Interestingly, all of our ASAP variants outperform the baseline and all other state-of-the-art methods, like BlurPooling \cite{zhang2019making}, ABlurPooling \cite{zou2020delving} or WaveletPooling \cite{li2020wavelet} on the corrupted data.
For all ResNet-like networks, the clean performance of the network is improved with our $\textrm{ASAP}_{\textcolor{orange}{\textrm{sp}}}$ \textcolor{orange}{and $\textrm{ASAP}_{\textcolor{orange}{\textrm{lp}}}$}. However, for MobileNet-v2 \cite{sandler2018mobilenetv2}, our \textcolor{orange}{downsampling} method\textcolor{orange}{s} cannot beat the baseline \textcolor{orange}{in terms of clean accuracy}.
We hypothesize that this behavior is due to the highly optimized training schedule used to train a MobileNet-v2.
Thus, including a new kind of downsampling might require additional finetuning of their training hyperparameters.
Analysing the adversarial robustness of our FLC Pooling and ASAP networks, we observe a trend towards higher robustness against FGSM and APGD for all downsampling methods, including the removal of high-frequency information in the frequency domain.
Hence, networks with ASAP and FLC Pooling can maintain high accuracy under FGSM attack.
The stronger APGD attack is able to fool the baseline in almost all cases completely. Other methods against aliasing are similarly weak in preventing the model from being fooled.
In contrary, networks using our ASAP and FLC Pooling cannot be fooled on all samples by the attack.
In summary, the improved robustness of models using FLC Pooling and ASAP indicates more stable feature learning.

\vspace{0.2cm}
\noindent\textbf{For low-resolution data}, we train ResNet-18 \cite{he2016deep} and Wide-ResNet-50-2 \cite{zagoruyko2016wide} (WRN-50-2) models on CIFAR-10 \cite{krizhevsky2009learning} with five different random seeds per network architecture.
We compare the standard baseline network, BlurPooling \cite{zhang2019making}, adaptive BlurPooling \cite{zou2020delving} (ABlurPooling) and WaveletPooling \cite{li2020wavelet}.
All networks are trained with the same set of hyperparameters: $150$ epochs, a batch size of $256$, a cosine learning rate schedule with a maximum learning rate of $0.2$ and a minimum of $0.0$, a momentum of $0.9$, and a weight decay of $0.002$.
We utilize label smoothing with a factor of $0.1$, and Stochastic Gradient Descent (SGD) for optimization.

\autoref{tab:cleantrain_cifar_acc} shows the results of the low-resolution ($32\times32$ pixel) dataset, CIFAR-10. FLC Pooling and ASAP consistently outperform the baseline in terms of robustness while maintaining similar clean performance indicating that stable representations have been learned. On ResNet-18, ABlurPooling \cite{zou2020delving} shows the highest robustness against adversarial attacks, yet with a slight decrease in clean performance and robustness against common corruptions.

\begin{figure*}[t]%
\textcolor{orange}{\tiny \qquad\qquad\qquad\quad\quad FGSM \cite{harnesssing} \qquad\quad\qquad\qquad\qquad\qquad\qquad\qquad\qquad\qquad PGD \cite{pgd} }\\
\centering
\includegraphics[width=0.8\textwidth]{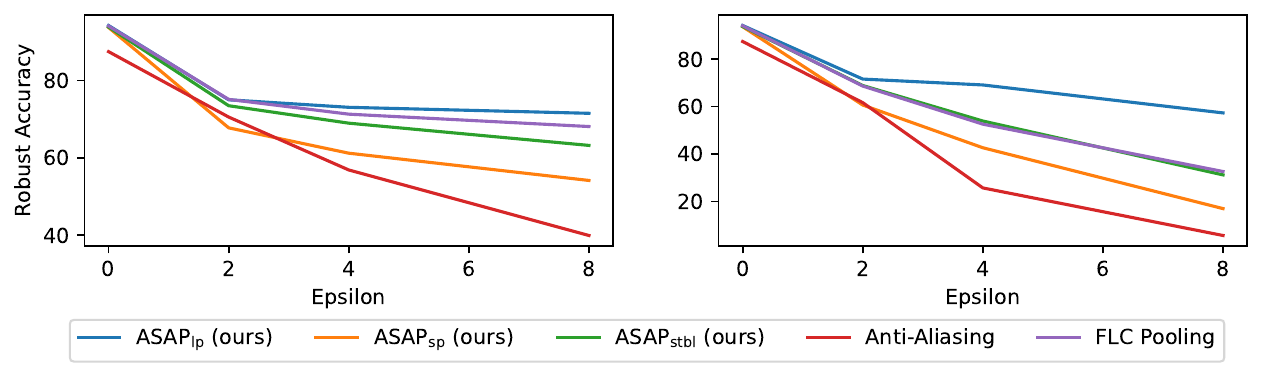}
\caption{\textcolor{orange}{Evaluationg different attack $\epsilon$ budgets of ResNet-50 on CIFAR-10 without adversarial training. We compare models using FLC Pooling and ASAP to the approach by \cite{rodriguezmunoz2022driver} (Anti-Aliasing). Our downsampling variants consistently exhibit higher robust accuracy on adversarial attacks than \cite{rodriguezmunoz2022driver}.}
}\label{fig:eps_native_abl}
\end{figure*}

Further, we compare to \cite{rodriguezmunoz2022driver} which use anti-aliasing mechanisms for downsampling and the activation function. Figure \ref{fig:eps_native_abl} presents this comparison under different $\epsilon$ budgets to examine the stability of the learned features. Our improved downsampling techniques can consistently provide more stable features leading to higher robustness compared to \cite{rodriguezmunoz2022driver} for all $\epsilon$ budgets without adversarial training.  

\begin{table}[t]
\scriptsize
  \caption{%
  Comparison of ResNet-50 models that are trained non-adversarially or adversarially with FGSM or PGD on ImageNet-1k.
  Accuracies \textcolor{orange}{(in percent)} shown on clean and perturbed (AutoAttack \cite{auto_attack}) validation images.
  We compare against models reported on RobustBench \cite{robust_bench}.}
\centering
    \begin{tabular}{cl|cc}
    \toprule
    \multicolumn{2}{c|}{Method} & Clean & \makecell{AA $L_{\inf}$\\
    $\epsilon = \frac{4}{255}$} \\
    \midrule
    \multicolumn{2}{c|}{Non-adversarial training \citeyear{robust_bench}} & 76.52 & 0.00 \\
    \midrule
         \multirow{3}{*}{\rotatebox[origin=c]{90}{FGSM}} & FLC Pooling (ours)  & 63.52 & 27.29 \\
         & $\text{ASAP}_{\textcolor{orange}{\textrm{sp}}}$ (ours)  &  64.51 & 30.93 \\
         & Wong et al., 2020 \citeyear{wong2020fast} & 55.62 & 26.24\\
         \midrule
         \multirow{3}{*}{\rotatebox[origin=c]{90}{PGD}} & Robustness lib, 2019 \citeyear{robustness_github} & 62.56 & 29.22\\
         & Salman et al., 2020 \citeyear{salman2020adversarially} & 64.02 & 34.96\\
         & $\textcolor{orange}{\text{ASAP}}_{\textcolor{orange}{\textrm{sp}}}$ \textcolor{orange}{(ours)}
         & 64.54 & 31.02 \\
         \bottomrule
    \end{tabular}
    \label{tab:imagenet_fgsm_trained}
\end{table}

\begin{figure*}[t]%
\centering
\includegraphics[width=0.8\textwidth]{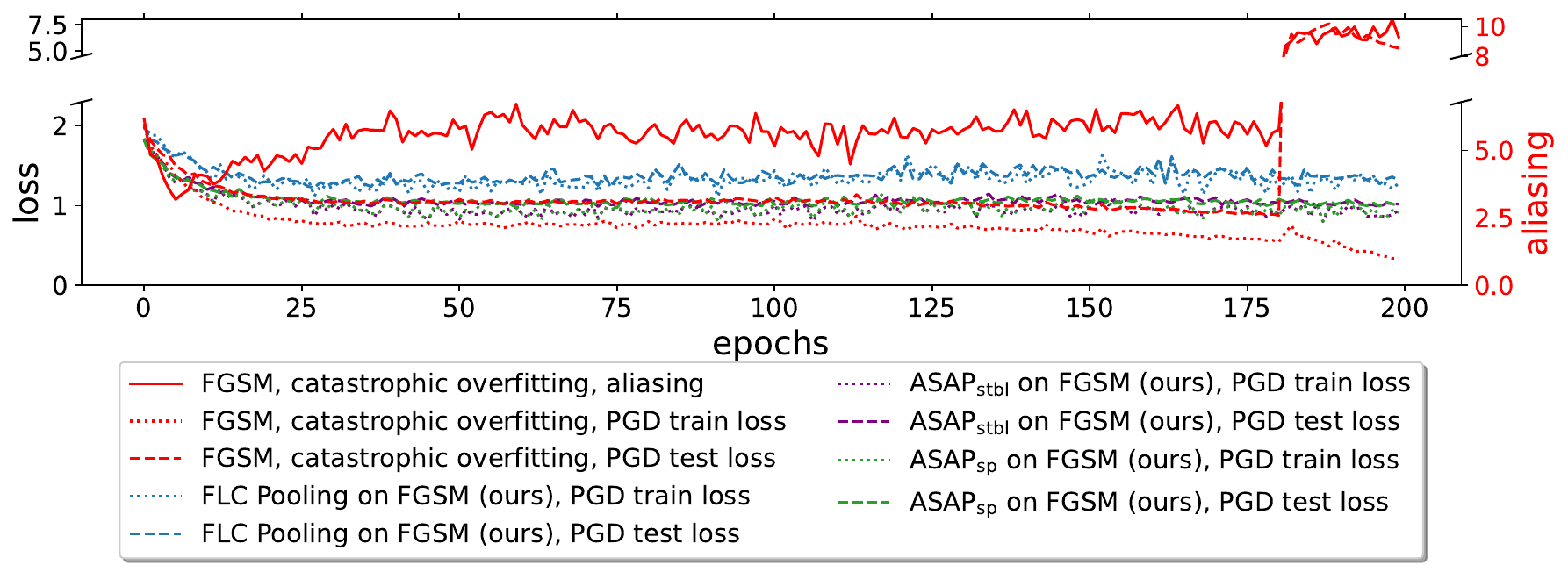}
\caption{Example of FGSM adversarial training facing catastrophic overfitting and its relationship to aliasing. FGSM training is prone to catastrophic overfitting (red lines) and experiences a huge increase in aliasing (red solid line) as soon as catastrophic overfitting happens, i.e. the test error on stronger adversaries like PGD increases (red dashed line) while the training error (red dotted line) stays low. Our methods, FLC Pooling and ASAP, are able to train with the fast FGSM adversarial training while preventing catastrophic overfitting (dashed and dotted lines).
}\label{fig:cata_ov}
\end{figure*}

\begin{figure}[t]%
\centering
\includegraphics[width=0.95\columnwidth]{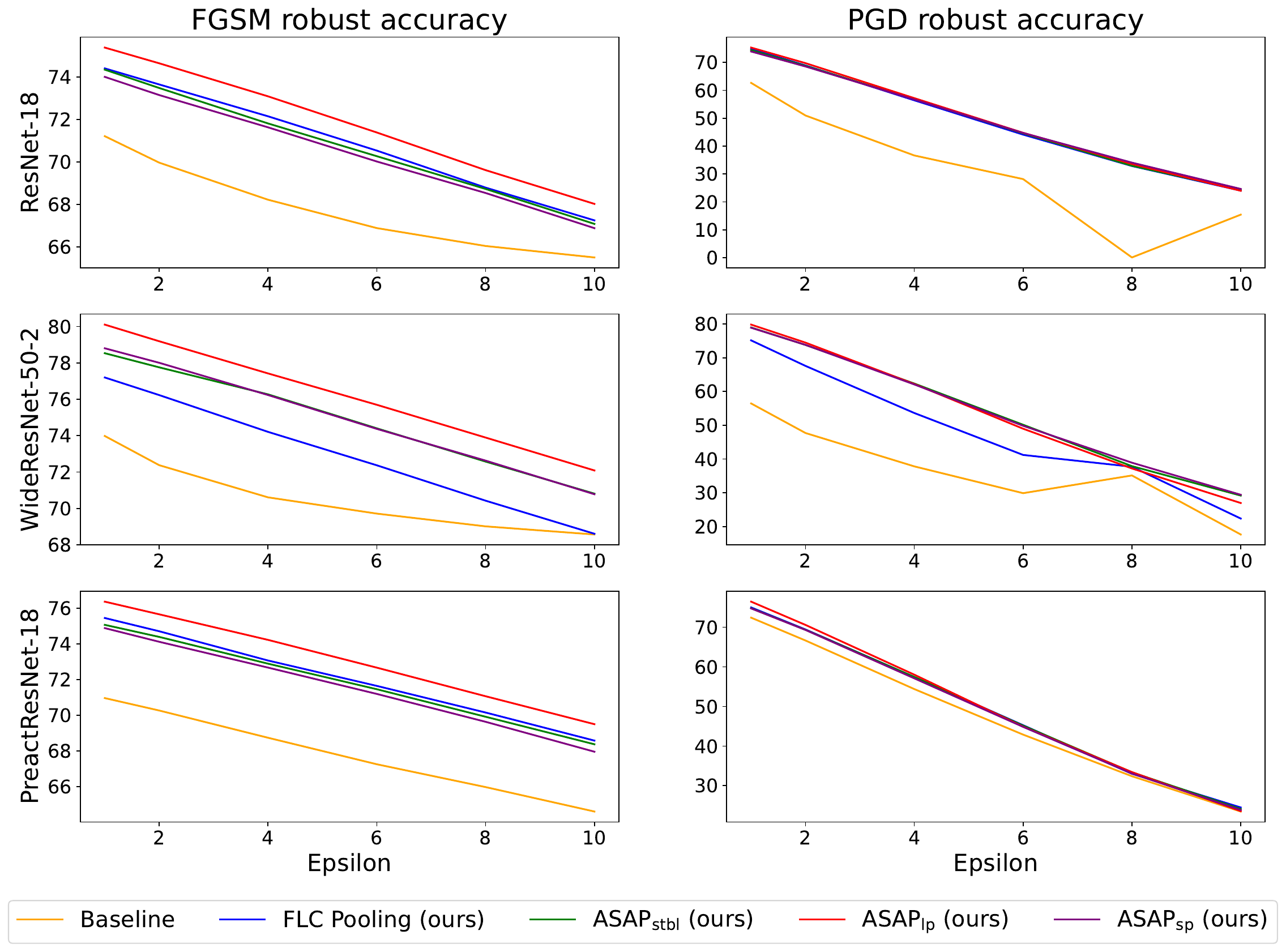}
\caption{Evaluation \textcolor{orange}{(accuracy in percent)} of networks adversarially trained with FGSM \cite{harnesssing} on \textcolor{orange}{CIFAR-10 evaluated on FGSM \cite{harnesssing}} (left) and PGD \cite{pgd} (right) adversaries with different $\epsilon$ budgets. Our three ASAP variants consistently exhibit higher robust accuracy on all architectures, adversarial attacks and \textcolor{orange}{across $\epsilon$ values} than the baseline.
}\label{fig:eps_fgsm}
\end{figure}

\subsection{Catastrophic Overfitting}
\label{subsec:cata_ov}
Catastrophic overfitting refers to the issue that models adversarially trained with FGSM \cite{harnesssing} tend to overfit to the FGSM attack \cite{kim2021understanding}. This usually only happens after several training epochs and leads to very low robustness towards other attacks such as PGD \cite{pgd}. In \cite{grabinski2022aliasing}, we observed that catastrophic overfitting in FGSM adversarial training often coincides with a high amount of aliasing in the model's downsampling layer. As a consequence, we assume that FLC Pooling and ASAP  should both reduce the risk of catastrophic overfitting in FGSM adversarial training, and thus facilitate to use this cheap training alternative for practically good results. 

\vspace{0.2cm}
\noindent\textbf{Qualitative Analysis.}
 \autoref{fig:cata_ov} confirms our hypothesis for the example of FGSM training on CIFAR-10 using PreAct-Resnet-18 (PRN-18), where FLC Pooling, \textcolor{orange}{$\textrm{ASAP}_{\textcolor{orange}{\textrm{stbl}}}$} and $\textrm{ASAP}_{\textcolor{orange}{\textrm{sp}}}$ \textcolor{orange}{which for now offered the most favorable trade-off between performance and efficiency,} exhibit a low PGD test error while the baseline experiences an increase in PGD test error simultaneously to an increase in aliasing.

\vspace{0.2cm}
\noindent\textbf{{Adversarial Training} on High-Resolution Data.} We train FLC \textcolor{orange}{Pooling} and $\textrm{ASAP}_{\textcolor{orange}{\textrm{sp}}}$ with FGSM \cite{wong2020fast} \textcolor{orange}{and PGD \cite{pgd}} adversarial training  on ImageNet-1k. 
We trained on a ResNet-50 and compare to models reported on RobustBench \cite{robust_bench} with the same architecture. \autoref{tab:imagenet_fgsm_trained} shows that our \textcolor{orange}{FGSM trained} model outperforms the baseline model trained with FGSM \cite{wong2020fast} in robust and clean accuracy, \textcolor{orange}{and even} the model trained by \cite{robustness_github} which takes \textcolor{orange}{significantly longer} than our method as shown in \autoref{tab:time_report}. The model by \cite{salman2020adversarially} achieves higher robustness, while being slightly worse on clean samples than our \textcolor{orange}{FGSM trained} $\textrm{ASAP}_{\textcolor{orange}{\textrm{sp}}}$ \textcolor{orange}{model}. However, this model uses the training by \cite{madry2017towards} which uses a multi-step adversarial attack, \textcolor{orange}{with extra data}. Since there is no release of the training script of this model on ImageNet, we can only roughly estimate their training times. As the training 
is based on PGD, we assume an increase in training time of \textcolor{orange}{at least a} factor \textcolor{orange}{of} six \textcolor{orange}{compared to our $\textrm{ASAP}_{{\textrm{sp}}}$ with FGSM}. 
\textcolor{orange}{Further, \autoref{tab:imagenet_fgsm_trained} also indicates that PGD training can benefit from proper downsampling. For this evaluation, we train a $\textrm{ASAP}_{{\textrm{sp}}}$ ResNet-50 with the training schedule by \cite{robustness_github} and achieve higher robustness and clean accuracy than their baseline. We also achieve higher robustness on clean images than \cite{salman2020adversarially} while not relying on extra data.}\\

\begin{table}[ht]
\caption[]{Accuracy \textcolor{orange}{(in percent)} for several \textcolor{orange}{common} models trained with FGSM adversarial training \cite{harnesssing} on CIFAR-10. We report adversarial robustness against FGSM \cite{harnesssing} and PGD \cite{pgd} with 50 attack iterations and 10 random restarts. Both attack have an $\epsilon$ budget of $8/255$. We clearly see, that our ASAP which neither suffers from aliasing nor from sinc artifacts, is also more robust in combination with adversarial training.
}
\footnotesize
\begin{tabular}{@{}l@{\,}l@{\,\,}l@{\,\,}|@{\,\,}c@{\,\,\,}c@{\,\,\,}c@{}}
\toprule
 &&Method     &   Acc@1 &  FGSM  &   PGD \\ 
\midrule
  \multirow{5}{*}{\rotatebox{90}{ResNet-18}}&& Baseline   & 78.85 $\pm$ 1.74  & 34.49 $\pm$ 2.68 & 21.14 $\pm$ 14.88 \\
&&FLC Pooling 
 &   79.77 $\pm$ 0.49& 34.37 $\pm$ 1.07& 32.23 $\pm$ 0.68  \\ 
 & &$\textrm{ASAP}_{\textcolor{orange}{\textrm{stbl}}}$ & 79.59 $\pm$ 0.64 & 35.13 $\pm$ 0.75& 32.65 $\pm$ 0.59\\ 
&&$\textrm{ASAP}_{\textcolor{orange}{\textrm{lp}}}$ & \textbf{80.63 $\pm$ 0.14} & \textbf{37.04 $\pm$ 0.65} & 33.43 $\pm$ 0.13 \\
&&$\textrm{ASAP}_{\textcolor{orange}{\textrm{sp}}}$ & 79.19 $\pm$ 0.32 & 35.44 $\pm$ 0.75 & \textbf{33.68 $\pm$ 0.31} \\
\midrule
\multirow{5}{*}{\rotatebox{90}{WRN-50-2}}& &Baseline &79.42 $\pm$ 0.34 & 39.18 $\pm$ 7.15&  23.36 $\pm$ 16.35 \\ 
&&FLC Pooling
 & 82.94 $\pm$ 0.89 & 39.23 $\pm$ 0.32 & 29.69 $\pm$ 11.81 \\ 
 & &$\textrm{ASAP}_{\textcolor{orange}{\textrm{stbl}}}$ & 83.63 $\pm$ 0.14 & \textbf{39.67 $\pm$ 0.28} & 37.62 $\pm$ 0.24 \\ 
&&$\textrm{ASAP}_{\textcolor{orange}{\textrm{lp}}}$ & \textbf{84.60 $\pm$ 0.13} & 39.56 $\pm$ 0.88 & 36.99 $\pm$ 0.19\\
&&$\textrm{ASAP}_{\textcolor{orange}{\textrm{sp}}}$  & 83.26 $\pm$ 0.24 & 39.16 $\pm$ 0.37 & \textbf{38.86 $\pm$ 0.17} \\
\midrule
 \multirow{5}{*}{\rotatebox{90}{PRN-18}} & &Baseline   &77.92 $\pm$ 0.19 & 31.74 $\pm$ 0.56& 32.52 $\pm$ 0.35 \\ 
&&FLC Pooling  
& 79.99 $\pm$ 0.09  & 36.39 $\pm$ 0.74 & 33.15 $\pm$ 0.19 \\ 
&&$\textrm{ASAP}_{\textcolor{orange}{\textrm{stbl}}}$ & 79.91 $\pm$ 0.17 & 36.25 $\pm$ 0.20& 33.20 $\pm$ 0.14 \\ 
&&$\textrm{ASAP}_{\textcolor{orange}{\textrm{lp}}}$& \textbf{81.29 $\pm$ 0.20} & \textbf{38.02 $\pm$ 0.85} & \textbf{33.48 $\pm$ 0.05} \\
&&$\textrm{ASAP}_{\textcolor{orange}{\textrm{sp}}}$ & 79.77 $\pm$ 0.20 & 35.88 $\pm$ 0.49 & 33.35 $\pm$ 0.19 \\
\bottomrule 
\end{tabular}
\label{tab:fgsmtrained_cifar}
\end{table}

\noindent\textbf{{Adversarial Training} on Low-Resolution Data.}
For CIFAR-10 we trained each model architecture with FGSM adversarial training using three different random seeds.
All hyperparameters were kept consistent across architectures and downsampling methods.
Each network underwent $300$ training epochs with a batch size of $512$ and a cycling learning rate schedule ranging from a maximum of $0.2$ to a minimum of $0.0$. The momentum was set to $0.9$, and weight decay was set to $0.0005$.
We employed CrossEntropyLoss as the loss function and utilized Stochastic Gradient Descent (SGD) as the optimizer.
The budget for the adversaries during training is $\epsilon = 8/255$.

The results in \autoref{tab:fgsmtrained_cifar} indicate that ASAP, similar to FLC Pooling, learns more stable features during adversarial training leading to higher robustness against adversarial attacks than the baseline.
Particularly when confronted with more complex adversaries like PGD \cite{pgd} with 50 attack iterations and 10 random restarts, our $\textrm{ASAP}_{\textcolor{orange}{\textrm{lp}}}$ consistently outperforms the baseline and FLC Pooling.
The high variance in performance on PGD samples for the ResNet-18 and Wide-ResNet-50-2 baseline indicate that some of the trained models lose all their robustness against PGD during FGSM adversarial training due to catastrophic overfitting. In contrast, our FLC Pooling and ASAP do not experience this issue due to the stable feature learning and maintain high robustness against strong and simple adversaries in all models. For PreAct-ResNet-18, which is commonly used for adversarial training \cite{gowal2021improving,rade2021helperbased,rebuffi2021fixing}, none of the networks experiences catastrophic overfitting. This is one aspect, \textcolor{orange}{of} why this network architecture might be used often for adversarial training. Still, our FLC Pooling and ASAP outperform the baseline on clean and perturbed samples.
Furthermore, \textcolor{orange}{all} ASAP \textcolor{orange}{variants} exhibit improved robustness against FGSM attacks and higher clean accuracy compared to the baseline (up to $5$\% improvement against FGSM attacks and up to $4$\% improvement on clean data). 
\textcolor{orange}{$\textrm{ASAP}_{\textcolor{orange}{\textrm{lp}}}$} improves the clean as well as the robust performance for the smaller models like ResNet-18 and PreAct-ResNet-18.
The larger Wide-ResNet-50-2 only benefits from the \textcolor{orange}{large} padding for clean accuracy.

\subsection{Ablations}
\label{sec:abl}
\begin{table*}[ht]
\caption[]{ \textcolor{orange}{Ablation on stabilization. We report accuracy (in percent) on clean samples, perturbed samples with FGSM~\cite{harnesssing} and PGD~\cite{pgd} as well as corrupted samples~\cite{hendrycks2019robustness} for different settings of ASAP on ResNet-18 \cite{he2016deep} trained without adversarial training on CIFAR-10. Attacks are done with $\epsilon=1/255$ and corruption performance is reported as mean over all severities.
}}
\begin{center}
\footnotesize
\begin{tabular}
{l@{\,\,\,}|@{\,\,\,}c@{\,\,\,\,}c@{\,\,\,\,}c@{\,\,\,\,}c}
\toprule
Architecture  &   Acc@1 &  FGSM  &   PGD & Corruptions\\
\midrule
Baseline &  93.03 $\pm$ 0.13 & 78.62 $\pm$ 0.28 & 72.49 $\pm$ 0.67  & 76.93  $\pm$ 0.45\\
$\mathrm{ASAP}_{\textrm{stbl}}$ & 93.12 $\pm$ 0.25 & 79.08 $\pm$ 0.43 & 75.06 $\pm$ 0.76 & \textbf{78.68 $\pm$ 0.19}\\
$\mathrm{ASAP}_{\textrm{lp}}$  & \textbf{93.24 $\pm$ 0.15} & \textbf{79.17 $\pm$ 0.23} & 74.94 $\pm$ 0.56 & 78.65 $\pm$ 0.33\\
$\mathrm{ASAP}_{\textrm{sp}}$ &  93.00 $\pm$ 0.12 & 79.12 $\pm$ 0.49 & 74.69 $\pm$ 1.36 & 78.42 $\pm$ 0.20\\
$\mathrm{ASAP}_{\textrm{lp}+\textrm{stbl}}$  & 93.22 $\pm$ 0.07 & 78.37 $\pm$ 0.77  & \textbf{75.65 $\pm$ 1.43} &   78.38 $\pm$ 0.38\\
$\mathrm{ASAP}_{\textrm{sp}+\textrm{stbl}}$ & 92.88  $\pm$ 0.11 & 77.29 $\pm$ 1.02 & 74.82 $\pm$    1.90 & 78.77 $\pm$ 0.17\\
\bottomrule              
\end{tabular}
\label{tab:stab_pad_abl}
\end{center}
\end{table*}
\begin{table*}[t]
\caption[]{Ablation on using different window functions in the frequency domain to reduce spectral artifacts. Mean clean and robust accuracy \textcolor{orange}{(in percent)} and their standard deviation on CIFAR-10 for five different window functions over five different random seeds. The best result is marked in \textbf{bold} and the second best via \underline{underlining}. Using no window function or a simple Gaussian window which even needs an additional hyperparameter performs quite poorly. All window functions which are a specialization of a Kaiser window perform reasonably well. Our Hamming window performs consistently well considering the top two performing methods.
}
\footnotesize
\begin{center}
\begin{tabular}{l|c|cccc}
\toprule
Window  &  Hyperparameter &   Acc@1 &  FGSM  &   PGD & Corruptions\\
\midrule
None & & 93.12 $\pm$ 0.19 & 78.92 $\pm$ 0.26 & 74.17 $\pm$ 0.60 &78.59 $\pm$ 0.29  \\
Hamming& & \underline{93.12 $\pm$ 0.25} & \textbf{79.08 $\pm$ 0.43} & \underline{75.06 $\pm$ 0.76} & \underline{78.68 $\pm$ 0.19} \\
Gaussian & $\sigma=(k-1)/6$ & 92.34 $\pm$ 0.15 & 77.85 $\pm$ 0.30 & 69.47 $\pm$ 0.40 & 78.53 $\pm$ 0.14\\
Hanning & & 93.13 $\pm$ 0.19 & \underline{79.23 $\pm$ 0.27} & \textbf{75.32 $\pm$ 0.92} & 78.56 $\pm$ 0.26\\
Kaiser & $\beta=7$ & \textbf{93.21 $\pm$ 0.17} &78.87 $\pm$ 0.33 & 74.57 $\pm$ 1.25 & 78.66 $\pm$ 0.21\\
Blackman &  &93.00 $\pm$ 0.17 & 78.82 $\pm$ 0.43 & 74.78 $\pm$ 1.16 & \textbf{78.77 $\pm$ 0.21} \\
\bottomrule              
\end{tabular}
\label{tab:window_ab}
\end{center}
\end{table*}

In the following, we conduct a series of ablations. First, we test the stability of our learned feature representations by examining the behavior of FLC Pooling and ASAP under attack with different $\epsilon$ values. 
Further, we investigate the effect of our different ASAP variants on the frequency spectrum of adversarial attacks.
Finally, we analyze the specific window functions to be used and the combination of the different ASAP variants.

\vspace{0.2cm}
\noindent\textbf{Ablating the Attack Strength.}
We assessed the behavior of FLC Pooling and ASAP under different budget settings of $\epsilon$. \autoref{fig:eps_fgsm} displays the mean robust accuracy trend across each architecture, varying the budget of $\epsilon$.
It is evident that ASAP consistently outperforms the baseline.
Moreover, $\textrm{ASAP}_{\textrm{lp}}$ demonstrates improved performance \textcolor{orange}{over all epsilon strengths under FGSM attack. In comparison, under PGD attack all our ASAP variants perform equally well, demonstrating improved stability in the learned feature representations.} 

\begin{figure}[t]%
\centering
\includegraphics[width=0.95\columnwidth]{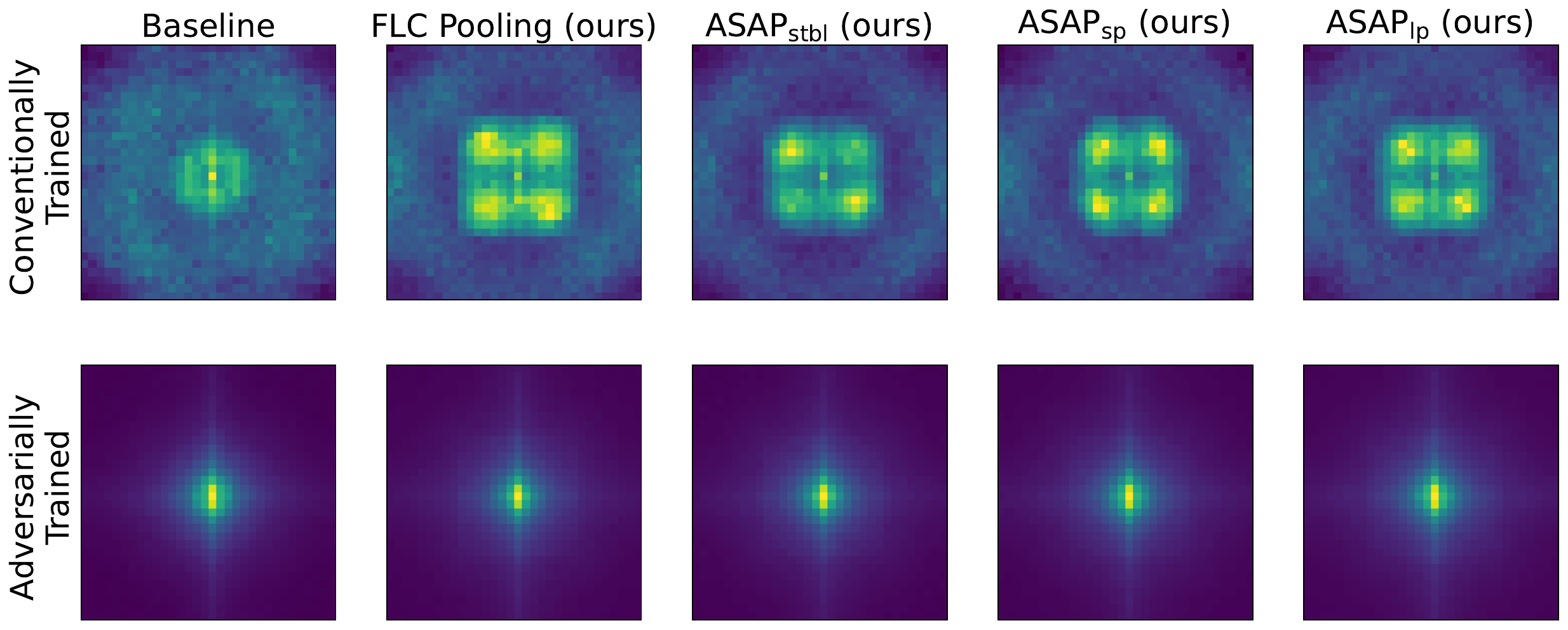}
\caption{Average difference in spectrum over 1000 CIFAR-10 images between the clean image and the attacked image with APGD \cite{croce2021mind}. For the conventionally trained networks (top row) the spectrum of the perturbation differ depending on the downsampling. However, for adversarially trained networks (bottom row) there is no difference clear difference. 
}\label{fig:adv_vs_base_trained}
\end{figure}
\vspace{0.2cm}
\noindent\textbf{Ablation on the Attack Spectrum. }
We investigate if there is a difference in perturbations created by \textcolor{orange}{APGD} \cite{auto_attack} depend\textcolor{orange}{ing} on the models' downsampling. \autoref{fig:adv_vs_base_trained} shows the perturbations created by \textcolor{orange}{APGD} on conventionally trained models (\textcolor{orange}{top}) and adversarially trained models (\textcolor{orange}{bottom}). The perturbations on conventionally trained models target all frequency bands as shown in the spectrum difference. While the attack mostly targets \textcolor{orange}{low}-frequency bands for the adversarially trained models, there is no clear difference between models including conventional downsampling or our downsampling methods.

\vspace{0.2cm}
\noindent\textbf{Ablation on the Window Function. }
For our work, we mainly focused on the Hamming window,
but there are several widely known window functions that could be used to reduce sinc artifacts.
Thus, we ablated four additional choices for the window function in our ASAP method.
Here, we additionally evaluate a standard Gaussian kernel, a Blackman window, a Hanning window, and a Kaiser window with $\beta = 7$.
Similar to the Kaiser kernel, we needed to choose an additional hyperparameter $\sigma$ for the Gaussian kernel.
We set sigma $\sigma$ in relation to the kernel size $k$ such that $\sigma = (k-1)/6$ as the length of $99$ percentile of the Gaussian pdf is $6\sigma$.
\autoref{tab:window_ab} presents the performance of five different random seeds trained on CIFAR-10 with the mentioned different window functions.
One can note that the models using a Gaussian window do not support the robustness of the network well,
while all models based on a Kaiser window \footnote{Hamming, Hanning and Blackman window are all specializations of a Kaiser window with fixed $\beta=6.0$, $\beta=5.0$ and $\beta=8.6$ respectively.} perform similarly well on clean, perturbed and corrupted data. When considering the top two performing methods, the Hamming window used for our ASAP performs consistently well and is thus a good choice.\\

\begin{table}[t]
\caption[]{Runtime of AT in seconds per epoch over 200 epochs and a batch size of 512 trained with a PreAct-ResNet-18 for training on the original CIFAR-10 dataset without additional data. Experiments are executed on one Nvidia Tesla V100. 
Evaluation for clean and robust accuracy, higher is better, on APGD \cite{croce2021mind} with our trained models. The models reported by the original authors may have varying results due to different hyperparameter selection. 
The top row reports the baseline without AT.}
\footnotesize
\begin{tabular}{@{}l@{\,}|@{\,\,}c@{\,\,}|@{\,\,}c@{\hspace{0.15cm}}c@{}}
\toprule
Method  &     Avg \#seconds & \multicolumn{2}{c}{Acc \textcolor{orange}{(\%)}}\\ 
 & per epoch  &  Clean &  APGD    \\ 
\midrule
Baseline & 6.6 $\pm$ 0.01 & 93.06 & 0.00 \\ 
\midrule
FGSM    \& early stopping \citeyear{wong2020fast} &   12.6 $\pm$ 0.01  & 82.88       & 11.82\\ 
FGSM    \& FLC Pooling (ours) &    14.7 $\pm$ 0.01   & 80.94 & 31.16  \\ 
FGSM \& $\textrm{ASAP}_{\textcolor{orange}{\textrm{stbl}}}$ (ours) & 15.6 $\pm$ 0.08  & 80.47 & 31.75 \\ 
FGSM \& $\textrm{ASAP}_{\textcolor{orange}{\textrm{sp}}}$(ours) &  17.1 $\pm$ 0.16  &  80.47 & 31.40\\
FGSM \& $\textrm{ASAP}_{\textcolor{orange}{\textrm{lp}}}$ (ours) &  36.4 $\pm$ 0.01 & 81.12 & 31.39\\ 
PGD \citeyear{pgd} &       115.4 $\pm$ 0.2  & 83.11 & 41.12 \\ 
Robustness lib \citeyear{robustness_github} &  117 $\pm$  19.0 &76.37 & 33.09 \\ 
AWP \citeyear{wu2020adversarial}&  179.4 $\pm$  0.4  &82.61 & \textbf{53.53} \\ 
MART \citeyear{Wang2020Improving} & 180.4 $\pm$ 0.8  &55.49 & 10.03 \\ 
TRADES \citeyear{trades} &     219.4 $\pm$ 0.5  & 81.49 & 49.65 \\ 
\bottomrule
\end{tabular}
\label{tab:time_report}
\end{table}

\noindent\textbf{\textcolor{orange}{Ablation on ASAP Variants. }}
\textcolor{orange}{Our method to stabilize the FFT by transposing (ASAP$_\textrm{stbl}$) can also be combined with padding. Table \ref{tab:stab_pad_abl} ablates on this combination (avg.~over five random seeds). The combination of both approaches, stabilization through transposing the signal and padding, yields no further systematic benefit, indicating that both approaches address the same issue.}

%
%
\section{Discussion on Efficiency }
\label{sec:padding_abliation}

Our FLC Pooling and ASAP stabilize feature learning leading to higher native robustness against common corruptions and adversarial attacks and reduce the risk of catastrophic overfitting during FGSM adversarial training.
Yet, we need to perform additional operations to transform from the spatial into the frequency domain and vice versa.
Hence, we achieve this increased feature stabilization with increased computational effort.
Especially when we are using large additional padding, the cost for the transformation becomes more expensive (e.g.~by a factor of $5.6$ in execution time).
However, we could show that small padding
can lead to an equal or better increase in robustness while increasing the computational effort only by a factor of $1.19$ compared to using no padding (\autoref{tab:pad_time}).
In comparison, additional data augmentation for feature stabilization to be more robust against common corruptions increases the number of samples that need to be learned, and adversarial training requires several forward and backward passes for each batch. Our analysis shows that adding additional FFT operations 
only increases the training time by a factor of $1.3$, while using more sophisticated adversarial training increases the training time by at least a factor of $9$ or even $17$ (\autoref{tab:time_report}).
Our proposed ASAP increases the training time dependent on the stabilization process used.
Simple stabilization by the transpose operation and small padding increases the adversarial training time per epoch by a factor of $1.2$. 
While large padding increases the training time by a factor of $2.8$ (\autoref{tab:time_report}), it is still faster than other sophisticated adversarial training methods.
Further, we want to point out that incorporating additional data like \textit{ddpm} \cite{ho2020denoising} which is a widely used source for adversarial training \cite{gowal2021improving,rade2021helperbased,rebuffi2021fixing} increases the training time by a factor of $20$.
In summary, adding FFT operations in the network achieves stable feature learning leading to high robustness while keeping the training time comparable low. 
\section{Conclusion}
We introduce two novel downsampling approaches in the Fourier domain. Both completely eliminate aliasing through aliasing-free pooling (FLC Pooling). Aliasing and Sinc Artifact-free Pooling (ASAP) additionally addresses sinc-interpolation artifacts.
Through extensive qualitative analysis, we motivate the benefit of artifact-free pooling.  Our quantitative analysis shows that FLC Pooling and ASAP learn beneficial, more stable features, leading, even without dedicated training, to improved robustness against common corruptions and adversarial attacks while maintaining high accuracy on clean data. Further, FLC Pooling and ASAP stabilize FGSM adversarial training and thus sustain high levels of robustness and accuracy by reducing the risk of catastrophic overfitting.

\backmatter




\newpage
\section*{Declarations}


\textbf{Acknowledgment} Steffen Jung and Margret Keuper acknowledge funding by the DFG Research Unit 5336 - Learning to
Sense.
\\
\noindent\textbf{Funding} Open Access funding enabled and organized by Projekt DEAL. 
\\
\noindent\textbf{Data Availability} The datasets worked with during the current study are publicly available at \href{https://image-net.org/index.php}{ImageNet-1k} and \href{https://www.cs.toronto.edu/~kriz/cifar.html}{CIFAR-10}, respectively.
\\
\noindent\textbf{Conflict of Interest} The authors declare that they have no conflict of interest.
\\
\noindent\textbf{Open Access} This article is licensed under a Creative Commons Attribution 4.0 International License, which permits use, sharing, adaptation, distribution and reproduction in any medium or format, as long as you give appropriate credit to the original author(s) and the source, provide a link to the Creative Commons licence, and indicate if changes were made. The images or other third party material in this article are included in the article’s Creative Commons licence, unless indicated otherwise in a credit line to the material. If material is not included in the article’s Creative Commons licence and your intended use is not permitted by statutory regulation or exceeds the permitted use, you will need to obtain permission directly from the copyright holder. To view a copy of this licence, visit http://creativecommons.org/licenses/by/4.0/.

\bibliography{main}

\end{document}